\documentclass{article}
\usepackage[final,dblblindworkshop]{neurips_2026}   %
\workshoptitle{Interpreting Agent Behavior (IAB)}
\usepackage[utf8]{inputenc}
\usepackage[T1]{fontenc}
\usepackage{hyperref}
\usepackage{url}
\usepackage{booktabs}
\usepackage{amsmath,amssymb}
\usepackage{graphicx}
\usepackage{xcolor}
\usepackage{microtype}
\usepackage{caption}
\usepackage{enumitem}
\usepackage{wrapfig}
\usepackage{placeins}   %

\graphicspath{{figures/paper/}}

\ifdefined\DRAFTSMALLN
  \usepackage{draftwatermark}
  \SetWatermarkText{DRAFT-smallN}
  \SetWatermarkScale{0.9}
  \SetWatermarkLightness{0.92}
\fi

\newcommand{\pa}{P1}\newcommand{\pb}{P2}\newcommand{\pc}{P3}\newcommand{\pd}{P4}
\newcommand{\pe}{P5}\newcommand{\pf}{P6}\newcommand{\pga}{P7a}\newcommand{\pgb}{P7b}\newcommand{\ph}{P8}

\newcommand{\RowsTotal}{2,406,697}
\newcommand{\LMRowsTotal}{1,191,747}

\newcommand{\BeliefCellsTotal}{180}
\newcommand{\BeliefCellsAdm}{123}

\newcommand{\SBCellsInformative}{46}
\newcommand{\SBCellsFloor}{8}
\newcommand{\SBCellsCeiling}{16}
\newcommand{\SBCellsMid}{22}

\newcommand{\SeatOnlyTotal}{14}
\newcommand{\CellsTotal}{60}

\newcommand{\ClSevenMemLadders}{15}

\newcommand{\ClSevenMemSup}{9}
\newcommand{\ClSevenMemRef}{4}
\newcommand{\ClSevenMemCon}{1}
\newcommand{\ClSevenMemInd}{1}

\newcommand{\ClEightMemSup}{4}

\newcommand{\ClMemZeroQwenGoofspielOne}{0.90}
\newcommand{\ClMemRungQwenGoofspielOne}{0.02}

\newcommand{\ClMemZeroQwenLeducOne}{0.11}
\newcommand{\ClMemRungQwenLeducOne}{0.10}

\newcommand{\ClMemZeroQwenDarkhexTwo}{0.90}
\newcommand{\ClMemRungQwenDarkhexTwo}{0.00}

\newcommand{\ClMemRungQwenGoofspielTwo}{0.32}

\newcommand{\ClMemZeroOlmoDarkhexOne}{0.53}
\newcommand{\ClMemRungOlmoDarkhexOne}{0.79}

\newcommand{\ClMemRungOlmoDarkhexTwo}{0.27}

\newcommand{\ClMemZeroOlmoKuhnTwo}{1.00}
\newcommand{\ClMemRungOlmoKuhnTwo}{0.29}

\newcommand{\ClMemZeroMistralLiarsdiceOne}{0.62}
\newcommand{\ClMemRungMistralLiarsdiceOne}{0.66}

\newcommand{\ClMemZeroMistralLeducTwo}{1.00}
\newcommand{\ClMemRungMistralLeducTwo}{0.17}
\newcommand{\ClMemZeroMistralLiarsdiceTwo}{0.63}
\newcommand{\ClMemRungMistralLiarsdiceTwo}{0.00}

\newcommand{\ClBeliefTotalOther}{24}

\newcommand{\ClBeliefWithheldPerProp}{19}
\newcommand{\ClBeliefStructural}{12}
\newcommand{\ClBeliefThin}{2}
\newcommand{\ClBeliefFailed}{5}

\newcommand{\ClPTwoCells}{30}
\newcommand{\ClPTwoIndet}{28}

\newcommand{\ClPSixBaseKuhn}{0.381}

\newcommand{\ClPSixBaseLeduc}{0.452}
\newcommand{\ClPSixQwenLiarsdice}{0.107}

\newcommand{\ClPSixBaseLiarsdice}{0.557}
\newcommand{\ClPSixOlmoKuhn}{0.321}

\newcommand{\ClPSixOlmoLeduc}{0.542}

\newcommand{\ClPSixOlmoLiarsdice}{0.860}

\newcommand{\ClPSixMistralKuhn}{0.516}

\newcommand{\ClPSixMistralLiarsdice}{0.158}

\newcommand{\ClPFiveGoofspielQwen}{$-0.009$}
\newcommand{\ClPFiveGoofspielOlmo}{$-0.067$}
\newcommand{\ClPFiveGoofspielMistral}{$-0.002$}

\newcommand{\ClLMNodesQwenMTwo}{796,543}
\newcommand{\ClBeliefFailQwenMTwo}{4}

\newcommand{\ClLMNodesMin}{381,237}
\newcommand{\ClLMNodesMax}{858,957}
\newcommand{\ClBeliefFailTotal}{4}
\newcommand{\ClActionFallbackTotal}{0}
\newcommand{\ClVEightPSevenaLo}{0.252}
\newcommand{\ClVEightPSevenaHi}{0.298}
\newcommand{\ClVEightPSevenaSame}{unchanged}
\newcommand{\ClVEightPSevenbLo}{0.402}
\newcommand{\ClVEightPSevenbHi}{0.535}

\newcommand{\ClSevenSBGoofspielLo}{0.50}
\newcommand{\ClSevenSBGoofspielHi}{1.00}
\newcommand{\ClVEightSBCeiling}{5}

\newcommand{\ClSevenSBOlmoKuhn}{1.00}

\newcommand{\ClPSevenaMeanSpread}{0.04}
\newcommand{\ClChanceQwen}{0.376}

\newcommand{\ClChanceOlmo}{0.361}

\newcommand{\ClChanceMistral}{0.375}

\newcommand{\ClChanceLo}{0.361}
\newcommand{\ClChanceHi}{0.376}
\newcommand{\ClGapSQwen}{0.18}
\newcommand{\ClGapSQwenLo}{0.17}
\newcommand{\ClGapSQwenHi}{0.19}
\newcommand{\ClGapSOlmo}{0.00}
\newcommand{\ClGapSOlmoLo}{0.00}
\newcommand{\ClGapSOlmoHi}{0.01}
\newcommand{\ClGapSMistral}{0.08}
\newcommand{\ClGapSMistralLo}{0.07}
\newcommand{\ClGapSMistralHi}{0.08}
\newcommand{\ClRespQwen}{0.19}
\newcommand{\ClRespQwenLo}{0.18}
\newcommand{\ClRespQwenHi}{0.21}
\newcommand{\ClRespQwenN}{12}
\newcommand{\ClRespOlmo}{$-$0.04}
\newcommand{\ClRespOlmoLo}{$-$0.04}
\newcommand{\ClRespOlmoHi}{$-$0.03}
\newcommand{\ClRespOlmoN}{4}
\newcommand{\ClRespMistral}{0.17}
\newcommand{\ClRespMistralLo}{0.15}
\newcommand{\ClRespMistralHi}{0.18}
\newcommand{\ClRespMistralN}{4}

\newcommand{\ClEqualValueKuhn}{0.80}
\newcommand{\ClChanceGameKuhn}{0.50}
\newcommand{\ClEqualValueLeduc}{0.87}
\newcommand{\ClChanceGameLeduc}{0.47}
\newcommand{\ClEqualValueGoofspiel}{0.51}
\newcommand{\ClChanceGameGoofspiel}{0.41}
\newcommand{\ClEqualValueLiarsdice}{0.56}
\newcommand{\ClChanceGameLiarsdice}{0.12}
\newcommand{\ClEqualValueDarkhex}{0.65}
\newcommand{\ClChanceGameDarkhex}{0.50}

\newcommand{\ClPSevenbQwenDarkhex}{0.35}

\newcommand{\ClRefSBKuhn}{0.00}

\newcommand{\ClRefSBLeduc}{0.00}

\newcommand{\ClRefSBGoofspiel}{0.43}

\newcommand{\ClRefSBLiarsdice}{0.07}

\newcommand{\ClRefSBDarkhex}{1.00}

\newcommand{\ClFourSBOlmoKuhn}{0.00--0.37}
\newcommand{\ClFourSBMistralKuhn}{1.00}

\newcommand{\RadarQwenMZeroPSevena}{0.75}

\newcommand{\RadarQwenMZeroPEight}{1.00}

\newcommand{\RadarWorseQwenMZero}{1}
\newcommand{\RadarCellsQwenMZero}{20}
\newcommand{\RadarOlmoMZeroPSevena}{0.72}

\newcommand{\RadarWorseOlmoMZero}{11}

\newcommand{\RadarMistralMZeroPSevena}{0.71}

\newcommand{\RadarMistralMZeroPEight}{0.43}
\newcommand{\RadarMistralMZeroPOne}{0.93}
\newcommand{\RadarMistralMZeroPFour}{0.98}

\newcommand{\RadarWorseMistralMZero}{6}

\newcommand{\RadarWorseQwenMOne}{7}

\newcommand{\RadarOlmoMOnePSevenb}{0.35}

\newcommand{\RadarWorseOlmoMOne}{12}

\newcommand{\RadarWorseMistralMOne}{7}

\newcommand{\RadarQwenMTwoPSevena}{0.70}
\newcommand{\RadarQwenMTwoPSevenb}{0.55}
\newcommand{\RadarQwenMTwoPEight}{0.39}

\newcommand{\RadarWorseQwenMTwo}{4}

\newcommand{\RadarOlmoMTwoPSevena}{0.66}

\newcommand{\RadarOlmoMTwoPEight}{1.00}

\newcommand{\RadarWorseOlmoMTwo}{9}

\newcommand{\RadarMistralMTwoPSevena}{0.64}

\newcommand{\RadarMistralMTwoPOne}{0.82}
\newcommand{\RadarMistralMTwoPFour}{0.89}

\newcommand{\RadarWorseMistralMTwo}{9}

\newcommand{\OptBaseStatedQwenKuhn}{0.18}

\newcommand{\OptBasePriorQwenKuhn}{0.05}

\newcommand{\OptBaseStatedQwenAll}{0.250}
\newcommand{\OptBaseUniformQwenAll}{0.353}

\newcommand{\OptBaseStatedOlmoKuhn}{0.23}

\newcommand{\OptBasePriorOlmoKuhn}{0.08}

\newcommand{\OptBasePriorOlmoAll}{0.180}

\newcommand{\OptBaseStatedMistralAll}{0.288}
\newcommand{\OptBaseUniformMistralAll}{0.342}
\newcommand{\OptBasePriorMistralAll}{0.167}
\newcommand{\OptBasePairs}{14}
\newcommand{\OptBaseBeatsUniform}{9}
\newcommand{\OptBaseNotBeatsUniform}{5}
\newcommand{\OptBaseNotBeatsUniformList}{Qwen's Dark Hex, OLMo's Liar's Dice, and Mistral's Leduc, Liar's Dice and Dark Hex}

\newcommand{\OptBaseBeatsPriorList}{OLMo's Goofspiel}

\newcommand{\OptTolFineMarginQwen}{0.18}

\newcommand{\OptTolCoarseMarginQwen}{0.15}

\newcommand{\OptTolFineMarginOlmo}{$-$0.01}

\newcommand{\OptTolCoarseMarginOlmo}{$-$0.01}

\newcommand{\OptTolFineMarginMistral}{0.07}

\newcommand{\OptTolCoarseMarginMistral}{0.10}

\newcommand{\OptPostStatedMarginQwen}{0.18}

\newcommand{\OptPostPosteriorMarginQwen}{0.19}

\newcommand{\OptPostStatedMarginOlmo}{$-$0.01}

\newcommand{\OptPostPosteriorMarginOlmo}{$-$0.03}

\newcommand{\OptPostStatedMarginMistral}{0.07}

\newcommand{\OptPostPosteriorMarginMistral}{0.10}

\newcommand{\OptUpdStatedQwenKuhn}{0.50}

\newcommand{\OptUpdPriorQwenKuhn}{0.50}

\newcommand{\OptUpdStatedQwenAll}{0.420}

\newcommand{\OptUpdPriorQwenAll}{0.491}

\newcommand{\OptUpdStatedMistralAll}{0.428}

\newcommand{\OptUpdPriorMistralAll}{0.417}

\newcommand{\AncFloorPSevenaMZeroLo}{0.34}
\newcommand{\AncFloorPSevenaMZeroHi}{0.35}
\newcommand{\AncPriorPSevenaLo}{0.17}
\newcommand{\AncPriorPSevenaHi}{0.23}
\newcommand{\AncPSevenaMZeroLo}{0.16}
\newcommand{\AncPSevenaMZeroHi}{0.29}

\newcommand{\AncPSevenbMZeroLo}{0.05}
\newcommand{\AncPSevenbMZeroHi}{0.24}

\newcommand{\AncAllPEightRef}{0.37}
\newcommand{\AncFloorPEightLo}{0.33}
\newcommand{\AncFloorPEightHi}{0.41}

\newcommand{\PSixSentinelLiarsdice}{0.30}
\newcommand{\AncCellsPThreeBelow}{39}
\newcommand{\AncCellsPThreeAbove}{100}
\newcommand{\AncCellsPThreeStraddle}{17}
\newcommand{\AncCellsPThreeDegenerate}{24}
\newcommand{\AncCellsPThreeTotal}{180}

\newcommand{\AncCellsPSixDegenerate}{22}

\newcommand{\AncCellsPSevenaTotal}{159}

\newcommand{\AncCellsPSevenbDegenerate}{5}

\newcommand{\AncCellsPEightDegenerate}{14}

\newcommand{\AncCellsBelow}{190}
\newcommand{\AncCellsAbove}{424}
\newcommand{\AncCellsStraddle}{31}
\newcommand{\AncCellsDegenerate}{65}
\newcommand{\AncCellsTotal}{710}
\newcommand{\AncCellsPSevenaPriorBelow}{144}
\newcommand{\AncCellsPSevenaPriorAbove}{15}

\newcommand{\AncBelowResolved}{3}

\newcommand{\AncAboveResolved}{26}
\newcommand{\AncBelowTiles}{OLMo at M1 on \pgb{}, Mistral at M1 on \pe{} and Mistral at M1 on \pga{}}
\newcommand{\AncUnresolvedTiles}{Qwen at M2 on \pf{}, Qwen at M2 on \ph{}, OLMo at M0 on \pc{}, OLMo at M1 on \pc{}, OLMo at M2 on \pc{}, OLMo at M2 on \pgb{}, Mistral at M0 on \pga{}, Mistral at M1 on \pgb{} and Mistral at M2 on \pc{}}
\newcommand{\AncUnresolvedN}{9}
\newcommand{\AncConflictTiles}{Mistral at M1 on \pe{} and Mistral at M1 on \pga{}}

\newcommand{\AncResolvedWidthMedian}{0.005}
\newcommand{\AncBelowGapMedian}{0.04}
\newcommand{\AncAboveGapMedian}{0.10}
\newcommand{\AncBelowWithinTwo}{48}

\newcommand{\PSixGameBelowN}{1}
\newcommand{\PSixGameBelow}{OLMo at M2 in Liar's Dice}
\newcommand{\PSixGameAboveN}{1}
\newcommand{\PSixGameAbove}{Qwen at M2 in Leduc}
\newcommand{\AncColoured}{52}

\newcommand{\ClNullBandQwen}{0.02}
\newcommand{\ClNullBandQwenKuhnMax}{0.10}

\newcommand{\ClPFourOlmoDarkhex}{0.04}
\newcommand{\ClNullBandOlmo}{0.01}

\newcommand{\ClNullBandOlmoDarkhexMax}{0.05}

\newcommand{\ClNullBandMistral}{0.00}

\newcommand{\ClNullBandMaxPair}{0.10}

\newcommand{\OrdTvQwenAll}{0.21}

\newcommand{\OrdAncQwenAll}{0.79}
\newcommand{\OrdTwinQwenAll}{0.01}
\newcommand{\OrdTwinMaxQwen}{0.10}
\newcommand{\OrdSameActionQwenAll}{0.82}
\newcommand{\OrdSamePositionQwenAll}{0.13}

\newcommand{\OrdNeitherOlmoLiarsdice}{0.76}

\newcommand{\OrdTvOlmoAll}{0.27}

\newcommand{\OrdAncOlmoAll}{0.73}
\newcommand{\OrdTwinOlmoAll}{0.01}

\newcommand{\OrdSameActionOlmoAll}{0.71}
\newcommand{\OrdSamePositionOlmoAll}{0.16}

\newcommand{\OrdPairedMistralLeduc}{177}

\newcommand{\OrdNeitherMistralLiarsdice}{0.90}

\newcommand{\OrdTvMistralAll}{0.19}

\newcommand{\OrdAncMistralAll}{0.81}
\newcommand{\OrdTwinMistralAll}{0.00}

\newcommand{\OrdSameActionMistralAll}{0.87}
\newcommand{\OrdSamePositionMistralAll}{0.02}

\newcommand{\OrdPairedSmallLo}{2}
\newcommand{\OrdPairedSmallHi}{12}

\newcommand{\PTwoFirstKuhnAlwaysbet}{0.14}
\newcommand{\PTwoRestKuhnAlwaysbet}{0.34}
\newcommand{\PTwoRefLeducCfrref}{0.16}

\newcommand{\PTwoRefLeducAlwaysbet}{0.07}

\newcommand{\PTwoFirstGoofspielAlwaysbet}{2.12}
\newcommand{\PTwoRestGoofspielAlwaysbet}{2.50}

\newcommand{\PTwoFirstLiarsdiceAlwaysbet}{0.64}
\newcommand{\PTwoRestLiarsdiceAlwaysbet}{0.84}

\newcommand{\PTwoCovGoofspiel}{0.07}
\newcommand{\PTwoCovLiarsdice}{0.00}

\newcommand{\PTwoRefLo}{$-$0.04}
\newcommand{\PTwoRefHi}{0.16}
\newcommand{\PTwoRefCells}{20}
\newcommand{\PTwoRefNearTol}{0.04}
\newcommand{\PTwoRefNear}{18}
\newcommand{\RefExploitDarkhex}{0.00012}
\newcommand{\RefExploitGoofspiel}{0.000074}
\newcommand{\RefExploitKuhn}{0.0044}
\newcommand{\RefExploitLeduc}{0.0020}
\newcommand{\RefExploitLiarsdice}{0.013}
\newcommand{\RefExploitMax}{0.013\%}
\newcommand{\ScoreboardCells}{360}
\newcommand{\RescoreMZeroCells}{120}
\newcommand{\RescoreCellsMoved}{6}
\newcommand{\RescoreBlockCellsChanged}{16}
\newcommand{\RescoreBasisCellsChanged}{9}
\newcommand{\RescoreNoteCellsChanged}{1}
\newcommand{\RescoreLaterCells}{240}
\newcommand{\RescoreLaterCellsMoved}{12}

\newcommand{\armQ}{FullnQwenTwoFiveOneFourbInstruct}
\newcommand{\armO}{FullnOlmoTwoOneOneTwoFourOneThreebInstruct}
\newcommand{\armM}{FullnMistralNemoInstructTwoFourZeroSeven}
\newcommand{\SBv}[2]{\csname SB#1#2\endcsname}
\newcommand{\SBmcv}[2]{\csname SBmc#1#2\endcsname}
\newcommand{\Blk}[3]{\csname Blocker#1#2#3\endcsname}
\newcommand{\Admv}[2]{\csname Adm#1#2\endcsname}
\newcommand{\Blkv}[2]{\csname Blk#1#2\endcsname}
\newcommand{\ECEv}[2]{\csname ECE#1#2\endcsname}
\newcommand{\Rowsv}[1]{\csname Rows#1\endcsname}
\newcommand{\PSevenAv}[2]{\csname PSevenA#1#2\endcsname}
\newcommand{\PSevenBv}[2]{\csname PSevenB#1#2\endcsname}
\newcommand{\PFiveMv}[2]{\csname PFiveM#1#2\endcsname}
\newcommand{\Verdictv}[2]{\csname Verdict#1#2\endcsname}
\newcommand{\Meanv}[2]{\csname Mean#1#2\endcsname}
\newcommand{\POnev}[2]{\csname POne#1#2\endcsname}
\newcommand{\PFourv}[2]{\csname PFour#1#2\endcsname}
\newcommand{\PFourheadv}[1]{\csname PFourhead#1\endcsname}
\newcommand{\SBCellsRefBlind}{8}
\newcommand{\SBCellsRefVaries}{38}
\newcommand{\SBCellsCeilingRefVaries}{10}
\newcommand{\SBCellsFloorRefVaries}{8}
\newcommand{\SBCellsMidRefVaries}{20}
\newcommand{\ClSevenMemFloor}{2}
\newcommand{\ClSevenMemTestable}{13}
\newcommand{\ClRefSBLiarsdiceQwenLo}{0.15}
\newcommand{\ClRefSBLiarsdiceQwenHi}{0.22}
\newcommand{\PSixv}[2]{\csname PSix#1#2\endcsname}
\newcommand{\AdmCellsv}[2]{\csname AdmCells#1#2\endcsname}
\newcommand{\SBcellv}[3]{\csname SBcell#1#2#3\endcsname}

\title{Instrumenting Failure: Scoring Language Model Agents at Every Decision Node in Imperfect-Information Games\thanks{Code and the pinned artifacts behind every number: \url{https://github.com/mkingopng/instrumenting-failure}.}}

\author{%
  Michael Kingston \\
  School of Computer Science and Engineering\\
  UNSW Sydney\\
  \texttt{m.kingston@student.unsw.edu.au} \\
  \And
  Sonit Singh \\
  School of Computer Science and Engineering\\
  UNSW Sydney\\
  \texttt{sonit.singh@unsw.edu.au} \\
  \And
  Alan Blair \\
  School of Computer Science and Engineering\\
  UNSW Sydney\\
  \texttt{a.blair@unsw.edu.au}
}

\begin{document}
\maketitle
\begin{abstract}
Agent evaluations report task success and rarely say what an agent did at each decision.
We score every decision node of three 12B to 14B instruct-tuned language models (LMs) playing five zero-sum imperfect-information games against four opponents on eight properties from decision and game theory, five of them against a CFR+ equilibrium reference.
At each node, the LM also states a belief about the opponent's hidden state in a separate call.
Without memory and at temperature 0, stated beliefs sit \Meanv{\armQ}{PSevena} to \Meanv{\armM}{PSevena} in total variation distance from the equilibrium posterior, and where the action responds to the state its agreement with that belief exceeds chance by \ClRespOlmo{} to \ClRespQwen{}.
The two calls frame the opponent differently, so the result is a behavioural inconsistency and makes no claim about internal beliefs.
Under one prompt and greedy decoding, the LM's action never varies with the state in \SBCellsCeilingRefVaries{} of the \SBCellsRefVaries{} cells where the reference varies its own.
Memory lowers the constant-action share in \ClSevenMemSup{} of \ClSevenMemTestable{} testable ladders at temperature 0; at temperature 0.2 the cells at 1 fall from \SBCellsCeiling{} to \ClVEightSBCeiling{} of the \SBCellsInformative{} informative cells, while the belief results hold.
A gate withholds, with a stated reason, any belief score the run cannot support, and seven of the eight non-zero defect rates are attributable to the instrument, a representation, or the LMs.
\end{abstract}
\section{Introduction}
\label{sec:intro}

Agentic systems are increasingly being deployed in decision-making roles such as clinical triage, financial operations and logistics.
In these domains, the state is only partially observed, dynamics are stochastic, actions have sequential consequence, other agents may exert adversarial pressure, and failure is costly.
Such decisions must be sound and, where optimality is defined, optimal; \citet{vonneumann1944theory} axiomatised the expected-utility theory that defines it.

Agentic frameworks routinely conflate three distinct operations: reasoning, deciding, and acting~\citep{gao2026actonomy}.
ReAct interleaves reasoning and action; the choice of action given the agent's belief is assumed to happen inside generation and is never represented as a step of its own~\citep{yao2023react}.
Agent evaluations then score task success, but cannot distinguish a sound decision from a lucky one.
The faithfulness literature cautions against reading chain-of-thought (CoT) to understand the decision.
CoT explanations can systematically misrepresent the true reason for a model's output and may be a post hoc justification and not a record of the reasoning~\citep{turpin2023language}.
What these evaluations leave unmeasured is belief--action consistency: whether the belief an agent states is consistent with its action.
Prior work audits belief--action consistency against a belief the agent is shown~\citep{gao2026auditing} or with a judge rating a written rationale~\citep{lin2026howfar}.
We score each decision node against two computable targets.
The first is the equilibrium posterior, the Bayes posterior over the hidden state if the opponent plays a CFR+ equilibrium, against which we score the stated belief whichever opponent is faced.
The second is the best response to that stated belief, with later play valued under the CFR+ policy, against which we score the action.
We read this behaviourally.
The LM states the belief in a separate call made after the action call, and that call never sees the action.
What we score is therefore consistency between two outputs of one agent, and we leave the explanation of any gap to the mechanistic stage (Section~\ref{sec:limits}).
If the stated belief were the agent's operative belief, a gap between it and the action would be a failure of the step that turns a belief into an action.
We do not test that antecedent, which is a faithfulness question outside this paper, and Section~\ref{sec:belief} describes the main alternative explanation.

We write \emph{LM} for a transformer-based language model, rather than LLM, because the 12B to 14B models scored here are not large.
An \emph{LM agent} is an LM placed in a game through a fixed interface: the prompts, the action parser and the serving setup.
Zero-sum imperfect-information games make this consistency measurable because the CFR+ reference and the equilibrium posterior are both computable there.
A win rate says nothing about what an agent did at each decision.
Under uncertainty a poor decision still wins its share of hands, and a result won on bad play is brittle.
The object we score is the LM agent's policy, logged one row per decision node with the agent's stated belief and the equilibrium posterior side by side.

Three open-weight instruct LMs play five games against four principled opponents.
We score each decision node on eight properties from decision theory and game theory, five of them against a CFR+ (counterfactual regret minimisation) equilibrium~\citep{zinkevich2007regret,tammelin2014cfrplus}.
They are invariance (\pa), adaptivity (\pb), randomisation (\pc), payoff sensitivity (\pd), calibration (\pe), sequential rationality (\pf), belief error (\pga), belief--action consistency (\pgb, against the belief the LM states) and value of information (\ph).
\pga{} and \pgb{} are the two estimands of one property.
\pa{} and \pd{} use paired probe runs, \pf{} uses driven off-path runs, and \pb{} reads only at memory levels M1 and M2.
Every property is computed for at least one model (Section~\ref{sec:admiss}).
Our contributions are:
\begin{enumerate}[leftmargin=*,itemsep=0pt,topsep=1pt]
\item \textbf{An instrument.} We score every logged decision node against computable targets and log the stated belief beside the equilibrium posterior in the same row (Sections~\ref{sec:instrument}, \ref{sec:admiss}).
\item \textbf{An admissibility gate.} The gate withholds a belief score, with a stated reason, when the scorer cannot compute it or the run cannot show that the model responds to the state. Of the eight non-zero defect rates, seven are attributable to the instrument, a representation, or the LMs; one stays unclassified (Section~\ref{sec:admiss}).
\item \textbf{A map of action constancy.} At temperature 0, over the \SBCellsRefVaries{} (model, game, opponent) cells where the CFR+ reference varies its action, the LM's action never varies with the state in \SBCellsCeilingRefVaries{}, always does in \SBCellsFloorRefVaries{}, and \SBCellsMidRefVaries{} lie between. In all \SBCellsInformative{} informative cells, the counts are \SBCellsCeiling{}, \SBCellsFloor{}, and \SBCellsMid{}. The map is a diagnostic: it shows where the belief and action metrics can be interpreted, and it depends on the model, the prompt and the decoding together (Sections~\ref{sec:stateblind}, \ref{sec:memory}).
\item \textbf{A behavioural belief--action result.} At M0 and temperature 0, the action puts \Meanv{\armO}{PSevenb} to \Meanv{\armQ}{PSevenb} of its probability on a best response to the stated belief, against a chance level of \ClChanceLo{} to \ClChanceHi{}. Where the action responds to the state, the margin over chance is \ClRespQwen{} for Qwen, \ClRespMistral{} for Mistral and \ClRespOlmo{} for OLMo (Section~\ref{sec:belief}). The stated beliefs are \Meanv{\armQ}{PSevena} to \Meanv{\armM}{PSevena} from the equilibrium posterior. Against floor agents scored through the same pipeline, the stated belief closes \AncPSevenaMZeroLo{} to \AncPSevenaMZeroHi{} of the gap from a uniform stater to the posterior. The action closes \AncPSevenbMZeroLo{} to \AncPSevenbMZeroHi{} of the gap from chance to full consistency (Appendix~\ref{app:grid}). The result does not make a causal claim, nor do we examine belief representation at an activation level, which is the next stage of this work.
\end{enumerate}

\section{Related work}
\label{sec:related}

\paragraph{Game-playing evaluations.}
Existing evaluations differ in how much of the computable optimum they use.
Some score mainly outcomes: GTBench's outcome-based relative advantage~\citep{duan2024gtbench}, the GTO Wizard Benchmark's AIVAT-adjusted winnings against a Nash-approximating agent~\citep{provost2026gtowizard}, and Game Reasoning Arena's outcomes for LMs over OpenSpiel games, the substrate used here, with keyword tagging of their reasoning~\citep{cipolina2025arena}.
Others score against the solver: \citet{lin2026howfar} score chips against RL baselines, CFR+ and Deep CFR and judge-rate action--reasoning consistency, calling the mismatch a knowing--doing gap, and PokerBench scores per-decision accuracy against game-theory-optimal solver labels~\citep{zhuang2025pokerbench}.
Suspicion-Agent~\citep{guo2024suspicion} has the agent state beliefs as free text, but scores it by chips won and does not check those beliefs against a posterior.
To our knowledge, no prior evaluation obtains a stated belief beside a computable posterior at the same node and scores the action against the best response to that stated belief.
The two nearest lines of work are \citet{lin2026howfar}'s knowing--doing gap, which uses an LM judge to rate a written rationale against the action taken, and \citet{gao2026auditing}'s belief--action deviations, measured against a belief layer the agent is shown (below).
We score the action against the computable best response to a belief the agent states itself, and score that belief against the equilibrium posterior; we decompose play into named per-node properties (Section~\ref{sec:instrument}) and withhold, with a stated reason, any belief score the run cannot support.
Most of our properties formalise a failure already documented in small, fully observable matrix games, one-shot or briefly repeated. GPT-4 predicts the opponent's alternating pattern yet fails to act on it~\citep{akata2025repeated}, and contextual framing shifts play under a fixed payoff structure~\citep{lore2024strategic}. Payoff-magnitude scaling can shift play where strategic equivalence says it must not, though inconsistently across models~\citep{huynh2025understanding}, and an LM randomises with bias where the equilibrium is mixed~\citep{vidler2025randomness}.
In that order, they become the per-node estimands belief error and belief--action consistency (\pga, \pgb), invariance (\pa), payoff sensitivity (\pd) and randomisation (\pc).

\paragraph{Agent frameworks and belief-conditioned agents.}
Deployed agent frameworks presume that deciding happens somewhere inside reasoning.
Chain-of-thought prompting treats a reasoning chain as progress toward an answer~\citep{wei2022chain}; ReAct interleaves thought and action with no step between them~\citep{yao2023react}; tree-search variants use the LM itself to value candidate states~\citep{yao2023tree,zhou2023lats}.
None of them decides against a belief as a step of its own. In ReAct, evaluated in part on interactive decision-making tasks, a thought only extends the context from which the LM generates its next action, and no step selects among options. The tree-search variants select among candidates, but by the LM's own valuation rather than against a belief.
Deciding is so rarely treated separately that \citet{gao2026actonomy} argue for it as a category of its own. In their LM-assisted coding of 3{,}455 behavioural sentences from 211 agent papers, deciding is eighth of ten actions by frequency, at 5.1\% against 20.2\% for reasoning (their Fig.~8a).
None of these four frameworks keeps an explicit belief over hidden state or chooses its action as a best response to a belief it states; our instrument scores both (Section~\ref{sec:instrument}).
Belief-conditioned agents are studied in planning and in social deduction.
Tru-POMDP~\citep{tang2025trupomdp} builds the belief of a POMDP planner from an LM's confidence-weighted hypotheses over hidden state, and the planner, rather than the LM, chooses each action by belief tree search.
\citet{gao2026auditing} audit belief-conditioned Werewolf agents against an external, hand-specified belief layer and report action--belief consistency of about 0.21 in their active-belief arm, the rate at which the action exactly matches the layer's top-ranked target.
They treat it as a lower bound on how much the agent uses the belief.
Neither scores a belief the LM agent states itself against a computable posterior, and neither uses an equilibrium.
Their 0.21 should not be set beside our best-response shares: theirs is an exact match with the top target of a belief the agent is shown. Ours is membership of a best-response set for a belief the agent states, which is easier to hit than one target, so a higher share would not mean a more consistent agent.

\paragraph{Descriptive, solver-scored and mechanistic interpretation.}
\citet{gao2026actonomy} ask how agent behaviour should be read and answer with vocabulary, a grounded-theory taxonomy of 10 actions over 46 sub-actions.
A taxonomy names what an agent did; it cannot say what the agent should have done, having no target to score against.
Per-node property scoring supplies that target where it is computable: it says where play departed from it, and by how much.
The nearest mechanistic work is \citet{sobotka2026broken}: linear probes recover latent game variables from hidden states in imperfect-information games, and steering those representations moves the action only partially, which is how they obtain the observation--belief and belief--action gaps.
Belief--action consistency is a behavioural counterpart of their belief--action gap, measured at every decision node without access to activations, and our future work (Section~\ref{sec:limits}) runs their probes beside it.
\section{Methodology}
\label{sec:instrument}

\subsection{Problem definition}
\label{sec:problem}

Formal decision theory takes a consistent decision to be one that maximises expected utility under subjective beliefs, updated by conditioning on evidence~\citep{savage1954foundations}.
Such decisions are invariant to how the problem is described~\citep{tversky1981framing}; in games they are randomised where equilibrium requires it~\citep{vonneumann1944theory} and rational at every information state~\citep{selten1975reexamination,kreps1982sequential}.
The eight properties of Section~\ref{sec:setup} operationalise them.

We formalise evaluation as measuring the policy the interface induces (Section~\ref{sec:intro}) against a computable reference.
A two-player zero-sum imperfect-information game presents the acting player, at each \emph{information set} $I$, with everything it can distinguish; the opponent's private state stays hidden.
The LM agent's policy $\sigma(\cdot \mid I)$ is its distribution over the legal actions at $I$; the reference $\sigma^{*}$ is the CFR+ average policy~\citep{zinkevich2007regret,tammelin2014cfrplus}, a provable equilibrium approximation whose exploitability is at most \RefExploitMax{} of the largest payoff in each game (Table~\ref{tab:games}).
We measure the distance from $\sigma$ to $\sigma^{*}$, the distance from the agent's stated belief to the equilibrium posterior (\pga), and whether the action best-responds to that stated belief (\pgb).
The two can fail independently, as a correct belief with an inconsistent action or a wrong belief acted on consistently, which is why they are separate estimands.
The equilibrium posterior is the Bayes posterior when the opponent plays $\sigma^{*}$.
$\sigma^{*}$ is one equilibrium, and five of the eight properties depend on it; invariance, adaptivity and payoff sensitivity (\pa, \pb, \pd) do not.
Section~\ref{sec:limits} states what that dependence costs.

We use games because they provide a computable reference at every decision while the uncertainty stays real: partial observation, chance moves, and an adversarial opponent~\citep{kuhn1951poker,zinkevich2007regret}.
The key feature is sequentiality.
The agent must carry a belief across nodes, updating it on every action and observation, so a mis-weighted update is carried forward unless later evidence corrects it.
And because payoffs are mostly terminal, an action is valued by its continuation: it shapes what the opponent can infer and the reach probabilities of later information sets.
A one-shot matrix game tests neither.

\subsection{Experimental setup}
\label{sec:setup}

\paragraph{Games and reference.}
The games and the CFR+ reference both come from OpenSpiel~\citep{lanctot2019openspiel}: the reference is its CFR+ solver run on the same game implementation the agent plays, so a gap between the agent and the reference cannot come from a mismatch between two encodings of the rules.
We play five two-player zero-sum imperfect-information games: Kuhn poker, iterated at the match level~\citep{kuhn1951poker}; Leduc poker~\citep{southey2005bayesbluff}; four-card Goofspiel; one-die Liar's Dice~\citep{ferguson1991liarsdice}; and $2\times2$ Dark Hex.
We chose the games so that their structures differ and, between them, every property can be measured.
Kuhn, the base case, has one hidden card.
Leduc adds betting rounds and a public card, which gives off-path nodes for sequential rationality (\pf).
Goofspiel has simultaneous moves and a mixed equilibrium (\pc).
Liar's Dice requires updating the posterior across a bidding sequence.
Dark Hex adds information acquisition at a cost, which makes value of information (\ph) measurable there and only there.
The games also differ in size, length, number of actions and chance structure, so we do not attribute a difference between games to any one of them.
Claims rest on within-game comparisons across opponents, models and memory rungs, and we report cross-game differences as observations (Appendix~\ref{app:games}).
Off-path nodes, which the CFR+ average policy never reaches, are scored against that policy mixed with a small uniform tremble, so every node is reachable and the reference there is fixed; the tremble follows \citet{selten1975reexamination} but does not compute a perfect equilibrium.

\paragraph{Opponents.}
Every LM agent plays four opponents.
The first is an always-bet rule and the second a frozen reinforcement-learning (RL) self-play agent.
The third plays the CFR+ reference, sampling each action from its mixed strategy, and the fourth a Bayesian best-responder that computes its posterior under a CFR+ model of its opponent.
In Goofspiel and Dark Hex the Bayesian opponent plays its fallback policy at every node, so in those two games it is a fallback agent.
Each opponent punishes different defects.

\paragraph{What is logged.}
A \emph{node} is one decision of an LM agent, and each node is one logged row (prompt, legal mask, action, retries, reference policy, reward, provenance) with two belief columns; Appendix~\ref{app:glossary} collects the terms.
The \emph{stated belief} $\hat b$ is the agent's stated distribution over the opponent's hidden state, from a second call at the same node.
The \emph{equilibrium posterior} $\mu$ is the Bayes posterior there when the opponent plays $\sigma^{*}$.
It is exact for the CFR+ opponent and a benchmark for the other three.
We ask for the belief in a separate call at the same node, issued after the action call.
It receives the same state and memory and never sees the action, so the two outputs share only the state and memory.

In decision theory the action is a best response to the belief; \pgb{} measures how often the LM's action is one to the belief it states.
Because the two come from separate calls, a mismatch cannot tell an action that ignores that belief from one that follows a different, unstated belief.
The best-response set in \pgb{} contains every action that ties for the best value, to within $10^{-6}$.
Its continuation is valued under the reference policy, so \pgb{} measures no exploitation of the opponent actually faced.
Two choices followed the first results: we corrected how the best-response set behind \pgb{} is computed and re-scored, which moved \RescoreCellsMoved{} of the \RescoreMZeroCells{} M0 scoreboard cells, all \pgb{}, without changing any cell's pass or fail under the submission's thresholds\footnote{The submitted version's \RescoreBlockCellsChanged{} of \ScoreboardCells{} included \RescoreBasisCellsChanged{} cells the renderer moved by changing basis and \RescoreNoteCellsChanged{} annotation change. The M1 and M2 arms were re-scored before the paper's root was built: \RescoreLaterCellsMoved{} of their \RescoreLaterCells{} cells moved, all \pgb{} in the same two games.}, and we added the temperature-0.2 collection after the temperature-0 results were seen (Section~\ref{sec:limits}).

\paragraph{Distance.}
Distances between distributions are Total-Variation Distances:
\begin{equation}
\mathrm{TVD}(p, q) \;=\; \tfrac{1}{2} \textstyle\sum_{x} \lvert p(x) - q(x) \rvert ,
\label{eq:tvd}
\end{equation}
half the $L_1$ distance, 0 for identical distributions and 1 for disjoint support.
KL divergence fails here in both directions: stated beliefs put mass on states the equilibrium posterior rules out, making $\mathrm{KL}(\hat b \,\|\, \mu)$ infinite, and sometimes give zero to states it allows, making $\mathrm{KL}(\mu \,\|\, \hat b)$ infinite~\citep[Ch.~14]{lattimore2020bandit}. Smoothing makes both finite but leaves the score dependent on the smoothing constant.
TVD is finite, symmetric and at most 1, so no single node can dominate a cell average, and it equals the probability mass the stated belief must move to match the posterior; mass on ruled-out states is also reported on its own, as the support-mismatch rate (Appendix~\ref{app:defects}).
Jensen--Shannon divergence is also finite and bounded, but is measured in information, not probability mass.

\paragraph{Properties.}
The instrument defines eight per-node properties from decision theory and game theory.
Table~\ref{tab:properties} gives each one its estimand, its reading rule and its anchors, and Appendix~\ref{app:grid} the result that grounds it.
A floor is a trivial agent: some floors are scored on the LM's own rows (a uniform stater, the chance level), and others come from a uniform-random player run through the same pipeline (\pc, \pf). The ceiling is the ideal Bayesian agent, defined per property from the equilibrium posterior, or the reference where Table~\ref{tab:properties} says so; it is not the Bayesian opponent of Section~\ref{sec:setup}, which falls back to a fixed policy in two games.
\pa{} and \pd{} are scored on separate paired probe runs, replayed hand by hand against the main arm.
\pf{} is scored at driven nodes: off-path nodes the LM reaches because each hand opens with scripted moves into a subgame the CFR+ policy almost never visits.
Its floor is the same TVD for a uniform-random player.
\pb{} is read from the M1 and M2 arms only. \pc{} is scored as regret against the CFR+ best-response value, an estimand that stays defined for a point-mass policy at temperature~0. \ph{} is defined for Dark Hex only and reads for two of the three models at M0 (Section~\ref{sec:admiss}).

\begin{table}[th]
\centering\footnotesize
\setlength{\tabcolsep}{4pt}
\caption{\textbf{The eight properties as per-node estimands.} $\sigma$: LM agent's policy; $\sigma^{*}$: CFR+ reference; $\sigma^{\mathrm{tr}}$: CFR+ policy with a small uniform tremble, used at off-path nodes; $\hat b$: stated belief; $\mu$: equilibrium posterior; $\mathrm{BR}$: best-response set.}
\begin{tabular}{@{}p{0.04\linewidth}p{0.22\linewidth}p{0.35\linewidth}p{0.27\linewidth}@{}}
\toprule
 & Property & Per-node estimand & Reading rule and anchors \\
\midrule
\pa{} & Invariance & $\mathrm{TVD}(\sigma,\ \sigma\circ\pi)$, paired per node, under a payoff-preserving relabelling $\pi$ & null band under relabelling; reordering in Appendix~\ref{app:optional} \\
\pb{} & Adaptivity & shrinkage of the best-response gap over hands & read at M1 and M2 only; delta below 0, with an interval that clears 0; no usable ceiling (Appendix~\ref{app:optional}) \\
\pc{} & Randomisation & regret of the action taken against the CFR+ best-response value, minus the uniform-random player's regret on the same cell & margin above 0, with an interval that clears the random player's (the floor); ceiling the reference \\
\pd{} & Payoff sensitivity & $\mathrm{TVD}$ under $u \mapsto au+b$, paired per node & as \pa{} \\
\pe{} & Calibration & per-pair calibration error of $\hat b$ minus a uniform-stating baseline & Student-$t$ interval over seeds against zero (the uniform stater is the floor) \\
\pf{} & Sequential Rationality & $\mathrm{TVD}(\sigma,\ \sigma^{\mathrm{tr}})$ at driven off-path nodes & read against the uniform-random player (floor); ceiling 0 \\
\pga{} & Belief Error & $\mathrm{TVD}(\hat b,\ \mu)$ & floor the uniform stater (no-update prior beside it); ceiling 0 \\
\pgb{} & Belief--Action Consistency (stated belief) & fraction of actions in $\mathrm{BR}(\hat b)$ & floor the per-cell chance level; ceiling 1 \\
\ph{} & Value of Information & captured value of information, Dark Hex only & floor the uniform-random capture rate; ceiling 1; not measured where no informative node is reached \\
\bottomrule
\end{tabular}

\label{tab:properties}
\end{table}

\paragraph{Admissibility.}
We report a score per \emph{cell}, one (game, opponent) pair per model.
The \emph{admissibility gate} is the rule that decides whether a cell's belief score may be taken as evidence about the model; each cell is either admissible, withheld with a stated reason, or not measured.
The gate withholds a belief cell (\pe, \pga, \pgb) when the scorer cannot compute it (under 30 assessable rows, a failed alignment, or a malformed posterior).
It also withholds the cell when the state-sensitivity check fails or cannot run.
The check groups a seat's nodes by their set of legal actions.
It fails if, in some group where the equilibrium posterior changes, the model plays one action and states one belief across all prompts: neither its action nor its belief responds to what it sees.
It cannot run if no group holds two different prompts.
The \emph{state-blind share} of a cell is the share of its nodes at which the action is the same across prompts that differ in state (Section~\ref{sec:stateblind}).
A \emph{seat-only} cell is one in which no seat saw two distinct prompts; it carries no claim.
A withheld cell keeps its value and carries one of three reasons.
First, the reason is structural when, against a deterministic opponent in a game without chance events, the model never sees two states within a seat.
Second, it is thin when there are too few distinct trajectories.
Third, it is failed when the check ran and the model did not pass it, which is a result about the model.
Every interval is a Student-$t$ interval over the ten seeds of a cell, or the five seeds of a probe run.
A cell whose interval straddles its floor is reported as indeterminate rather than above or below it, since its seeds cannot then tell the LM from the trivial agent.
Beside each belief score the cell prints five measured defect rates.
They are null belief, zero posterior, key-set mismatch (the hidden-state keys differ), support mismatch, and identity (belief byte-equal to the posterior, a leakage check).
The defect rates inform the reading and never gate; the reading is a rule over measured quantities, so no hand-kept list of known problems can go stale.

\paragraph{Models, serving and arms.}
This paper documents the first part of a staged experiment: building and validating the instrument before doing the primary experiments. 
Small models, shallow games and modest hand and seed counts allow us to iterate quickly and validate the instrument end-to-end before compute goes to the headline tier, and we scope the claims to the 12B to 14B instruct tier (Section~\ref{sec:limits}).
We run three open-weight instruct LMs of one size tier: Qwen2.5-14B-Instruct, OLMo-2-13B-Instruct and Mistral-Nemo-Instruct-2407 (12B); Qwen, OLMo and Mistral below.
The interface is fixed: we serve each LM through vLLM at a pinned revision at temperature~0, with identical prompts and a belief call at every decision node.
An \emph{arm} is one LM agent over all five games and four opponents at one memory level and temperature.
At M0 an arm has ten seeds per cell and 1000 hands per seed.
We draw seats (who moves first) at random per hand.
The three M0 arms hold \RowsTotal{} rows, \LMRowsTotal{} of them LM decisions.
Probe runs (\pa, \pd) replay 100 hands per seed, paired hand by hand to the main arm through a separate probe pipeline.
For \pf, one driven run per model and game plays 60 hands per seed, each opening with the scripted moves and then played on by the LM against CFR+ from the seat that moves next. The Goofspiel and Dark Hex runs are withheld, since the off-path reference there is uniform (Appendix~\ref{app:grid}).
Two further arms per model add memory to the experiments.
M1 adds a running summary line, and M2 adds the summary plus a transcript of recent hands.
Each has 2000 hands per seed and ten seeds.
A second collection repeats the nine arms at temperature 0.2 (Appendix~\ref{app:temperature}).

\section{Results}
\label{sec:results}

\subsection{A structured share of decisions is blind to the state}
\label{sec:stateblind}
How often the LM's action ignores the state depends on the game and the model together.
The question is whether the agent's action changes when the state does.
For each legal-action set (the \emph{mask}: the actions legal at a node) and seat, the scorer collects the distinct prompts the agent saw. If one action covers every prompt in that group, the action did not respond to the state.
We count the group's nodes as state-blind. The \emph{state-blind share} of a cell is the proportion of its nodes counted this way. It runs from 0 (in every group the action changed with the prompt) to 1 (in no group did it).
Values in between mean the agent responds to the state at some nodes and ignores it at others.
Each node is classified as blind or not.
The share therefore measures how much of an agent's play ignores the state and leaves open whether the agent as a whole is state-blind.
Only groups in which one seat saw two or more distinct prompts enter the share, since only they separate blindness from lack of exposure.
We count prompts; information sets that render to the same prompt look identical to the agent, so they form one group.

\begin{figure}[!ht]
\centering
\includegraphics[width=300pt]{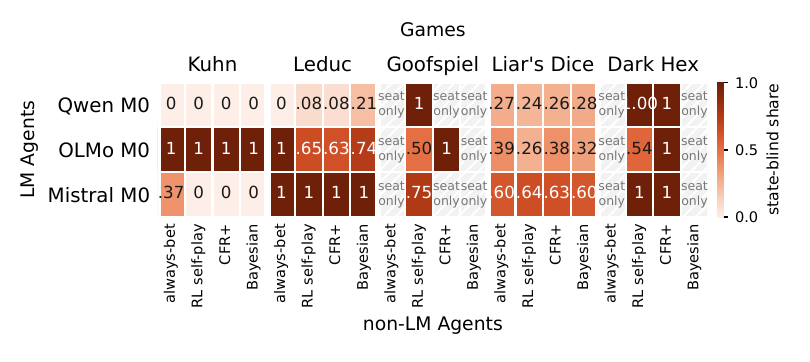}
\caption{\textbf{State-blindness has a game and model structure at temperature 0.} Share per (model, game, opponent) cell at M0 (1000 hands $\times$ 10 seeds), over nodes whose mask saw two or more prompts within one seat; hatched cells are seat-only and carry no claim. Figure~\ref{fig:stateblind-all} (Appendix~\ref{app:grid}) adds the M1 and M2 rows.}
\label{fig:stateblind}
\end{figure}

A cell is \emph{seat-only} when each seat saw just one prompt per mask, so the state never changed within a seat. It carries no claim (\SeatOnlyTotal{} of \CellsTotal, hatched in Fig.~\ref{fig:stateblind}).
Seat-only cells arise in the two games without chance events (Goofspiel and Dark Hex) when the opponent plays the same way every hand.
Greedy decoding at temperature~0 then repeats the model's most likely reply, so the model sees nearly the same hand each time and only its seat varies.

Figure~\ref{fig:stateblind} holds one tile per (model, game, opponent) cell.
The poker games already separate the models.
Qwen's state-blind share is 0 in Kuhn and at most \SBcellv{\armQ}{Leduc}{Bayesian} in Leduc.
Mistral's is 1 in Leduc; in Kuhn it is 0 against three opponents and \SBcellv{\armM}{Kuhn}{AlwaysBet} against always-bet.
OLMo's is high in both (\SBv{\armO}{Kuhn} in Kuhn, \SBv{\armO}{Leduc} in Leduc).
Liar's Dice sits at intermediate shares for every model (Qwen \SBv{\armQ}{Liarsdice}, OLMo \SBv{\armO}{Liarsdice}, Mistral \SBv{\armM}{Liarsdice}) and varies little by opponent.
In Goofspiel and Dark Hex the informative cells are at or near 1, with three exceptions.
They are Mistral and OLMo against RL self-play in Goofspiel (\SBcellv{\armM}{Goofspiel}{Rl}, \SBcellv{\armO}{Goofspiel}{Rl}) and OLMo against RL self-play in Dark Hex (\SBcellv{\armO}{Darkhex}{Rl}).
The reference itself is state-blind in some of these cells.
In Dark Hex its share is \ClRefSBDarkhex{} in every informative cell.
In Goofspiel the reference is state-blind in OLMo's two informative cells and in none of Qwen's or Mistral's; weighted by rows, that is \ClRefSBGoofspiel{} of the game's multistate rows.
It is \ClRefSBKuhn{} in Kuhn, \ClRefSBLeduc{} in Leduc and \ClRefSBLiarsdice{} in Liar's Dice pooled over rows, though in Qwen's four Liar's Dice cells it is \ClRefSBLiarsdiceQwenLo{} to \ClRefSBLiarsdiceQwenHi{} against Qwen's \SBv{\armQ}{Liarsdice}.
So constancy in Kuhn and Leduc is the LM's own; in Dark Hex and OLMo's Goofspiel cells it matches the reference, and in Qwen's Liar's Dice cells it partly does. Table~\ref{tab:refbaseline} lists both shares for every informative cell.
Excluding the \SBCellsRefBlind{} cells where the reference is state-blind leaves \SBCellsRefVaries{}.
There the LM's action never varies with the state in \SBCellsCeilingRefVaries{} cells, always does in \SBCellsFloorRefVaries{}, and \SBCellsMidRefVaries{} lie between.
Over all \SBCellsInformative{} informative cells the counts are \SBCellsCeiling{}, \SBCellsFloor{} and \SBCellsMid{}.
The shares sit at or near the floor or the ceiling in most cells.
Yet \SBCellsMid{} cells take intermediate values, so an agent can respond to the state at some nodes and ignore it at others.
The intermediate cells are Liar's Dice for every model, Leduc for Qwen and OLMo, and four single cells; a small game gives a cell few (mask, seat) groups, so its share can take only a few distinct values.
The share names the behaviour, but does not name its cause; the M0 reordering probe separates a positional bias from a label preference (Appendix~\ref{app:optional}).
At temperature 0.2 the number of cells at 1 falls from \SBCellsCeiling{} to \ClVEightSBCeiling{}.
Under the earlier collection reported in the workshop submission (prompt version 4), OLMo's share on Kuhn was \ClFourSBOlmoKuhn{} against \ClSevenSBOlmoKuhn{} here, and Mistral's was \ClFourSBMistralKuhn{} against \SBv{\armM}{Kuhn} (Appendix~\ref{app:temperature}).

\subsection{Memory lowers the constant-action share in most ladders}
\label{sec:memory}
\begin{figure}[!ht]
\centering
\includegraphics[width=280pt]{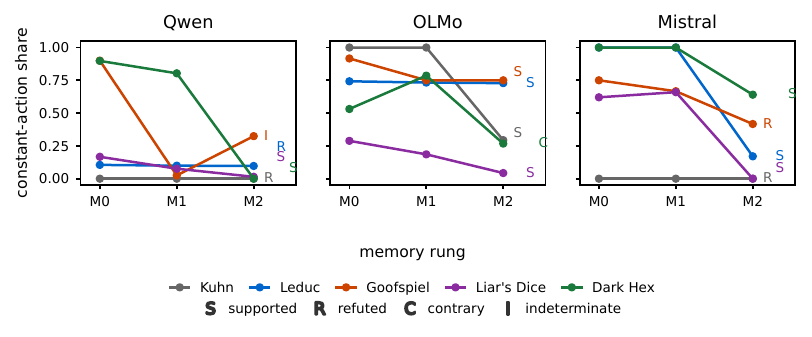}
\caption{\textbf{Memory lowers the constant-action share in \ClSevenMemSup{} of the \ClSevenMemTestable{} testable ladders and raises it marginally in \ClSevenMemCon{}.} Constant-action share per (model, game) memory ladder, pooled over seats and so not the within-seat share of Fig.~\ref{fig:stateblind}; M0 $\to$ M1 $\to$ M2, over the (legal-action set, seat) bins populated in both arms (M0 1000 and rungs 2000 hands $\times$ 10 seeds); letters are the ladder verdicts of Appendix~\ref{app:ladders} (S supported, R refuted, C contrary, I indeterminate).}
\label{fig:memory}
\end{figure}

Memory lowers the constant-action share in \ClSevenMemSup{} of the \ClSevenMemTestable{} testable ladders and raises it marginally in one; over all \ClSevenMemLadders{} ladders the read is indeterminate.
Without cross-hand state a model faces byte-identical prompts hand after hand, so a constant action alone cannot separate a policy from a reflex.
The memory rungs supply the contrast.
The constant-action share is the share of nodes whose action is the same across all hands that offer the same legal actions, pooled over seats.
If memory is used, the constant-action share should fall.
That share pools the seats the model alternated between, so it is not the within-seat share of Section~\ref{sec:stateblind} and its levels do not compare with Figure~\ref{fig:stateblind}'s.
The finding is the structure of the ladders (Fig.~\ref{fig:memory}).
Across \ClSevenMemLadders{} ladders (one model and game read at M0, M1 and M2) the read is indeterminate: \ClSevenMemSup{} supported, \ClSevenMemRef{} refuted, \ClSevenMemCon{} contrary, \ClSevenMemInd{} indeterminate.
Of the refuted ladders, \ClSevenMemFloor{} start at the floor (an M0 share no larger than the between-seed spread) and could not fall, so \ClSevenMemTestable{} ladders are testable and \ClSevenMemSup{} of those are supported.
A rung counts only where its change clears the ladder's between-seed spread; Appendix~\ref{app:ladders} gives the rules and every ladder.

Where the share starts at the floor, memory cannot lower it: Qwen's and Mistral's Kuhn read refuted, and Qwen's Leduc starts near it (\ClMemZeroQwenLeducOne{} to \ClMemRungQwenLeducOne{}) and does not move.
Mistral's Goofspiel is refuted for a different reason: its M1 change lies inside the between-seed spread and its M2 change equals it, so neither rung clears the spread (Table~\ref{tab:ladders}).
Where it starts at or near the ceiling, the large falls come at M2.
They are OLMo's Kuhn (\ClMemZeroOlmoKuhnTwo{} to \ClMemRungOlmoKuhnTwo{}), Mistral's Leduc (\ClMemZeroMistralLeducTwo{} to \ClMemRungMistralLeducTwo{}), Qwen's Dark Hex (\ClMemZeroQwenDarkhexTwo{} to \ClMemRungQwenDarkhexTwo{}) and Mistral's Liar's Dice (\ClMemZeroMistralLiarsdiceTwo{} to \ClMemRungMistralLiarsdiceTwo{}).
Qwen's Goofspiel falls at M1 (\ClMemZeroQwenGoofspielOne{} to \ClMemRungQwenGoofspielOne{}) and returns part of the way at M2 (\ClMemRungQwenGoofspielTwo{}), so its ladder is indeterminate.
One ladder rises marginally: OLMo's Dark Hex at M1 (\ClMemZeroOlmoDarkhexOne{} to \ClMemRungOlmoDarkhexOne{}) clears its between-seed spread by less than 0.01 and then falls at M2 (\ClMemRungOlmoDarkhexTwo{}); one supported ladder, OLMo's Leduc, clears its spread by the same margin (the daggers in Table~\ref{tab:ladders}). Mistral's Liar's Dice at M1 rises slightly (\ClMemZeroMistralLiarsdiceOne{} to \ClMemRungMistralLiarsdiceOne{}), inside the spread. We record \ClActionFallbackTotal{} action fallbacks in any arm, so the rise is not a fallback artifact.
The ladders show where the constant-action share moved and leave the cause open; adaptivity (\pb) is read separately (Section~\ref{sec:admiss}).

\subsection{Which belief cells are admissible}
\label{sec:admiss}
The gate admits every belief cell in Kuhn, Leduc and Liar's Dice and withholds \ClBeliefWithheldPerProp{} of \ClBeliefTotalOther{} cells per property in Goofspiel and Dark Hex, each with a stated reason.
A per-node metric can fail without announcing it: a scorer reads a malformed belief or a misaligned key set, and the printed number measures the defect.
The gate withholds a cell when the scorer cannot score it or the state-sensitivity check fails or cannot run, and prints the defect rates every admitted number still carries (Appendix~\ref{app:defects}).

\begin{figure}[!t]
\centering
\includegraphics[width=0.86\textwidth]{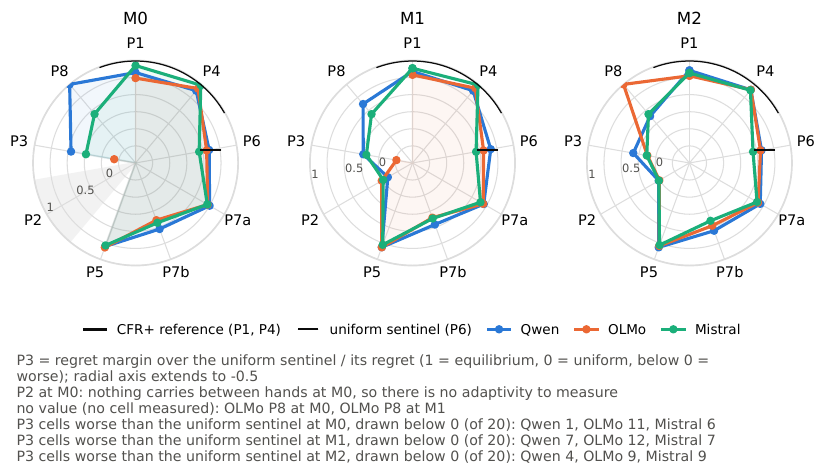}
\caption{\textbf{Per-property readings at M0, M1 and M2.} One panel per memory level on one radial scale. Each axis maps its own estimand to $[0,1]$ (\pa/\pd/\pf/\pga: $1-$TVD; \pgb: fraction; \pe: $1-$per-pair error; \pc: regret margin over the uniform-random player, drawn below 0 where worse; \ph: captured value of information); axes are not commensurate. Grey: not measured (\pb{} reads at M1 and M2 only); tick on \pf: the uniform-random player. The rim arc marks the CFR+ reference, at 1 on \pa{} and \pd{} by construction; its other axes are undefined.}
\label{fig:radar}
\end{figure}

The withheld cells are \ClBeliefStructural{} structural, \ClBeliefThin{} thin and \ClBeliefFailed{} failed, leaving \BeliefCellsAdm{} of \BeliefCellsTotal{} belief scores admitted in all.
The \ClBeliefFailed{} failed cells are a result about the model's stated belief: the reference's most likely action is also constant in each of them, so the fail reflects the unchanging belief, not the constant action.
They are Qwen, OLMo and Mistral against CFR+ in Dark Hex, OLMo against RL self-play in Dark Hex, and OLMo against RL self-play in Goofspiel.
Each played one action across prompts it could tell apart while stating one belief throughout.
\pb{} reads at M1 and M2 over two hand-index bins; \ClPTwoIndet{} of the \ClPTwoCells{} (model, game, rung) cells read indeterminate.
The others are both in Dark Hex with zero between-seed variance, one policy repeated, so neither rests on an interval and all \ClPTwoCells{} read as indeterminate here.
\pc{} reads in (game, opponent) cells, 20 per arm, each against the uniform-random player's own interval (Appendix~\ref{app:grid}).
Of its \AncCellsPThreeTotal{} cells over the nine arms, \AncCellsPThreeBelow{} lie below the random player with an interval that clears it, \AncCellsPThreeAbove{} above, \AncCellsPThreeStraddle{} straddle it and \AncCellsPThreeDegenerate{} are one policy repeated (Table~\ref{tab:cells}).
\ph{} reads in Dark Hex for Qwen and Mistral.
\ph{} is read against the uniform-random capture rate at the same nodes, \AncFloorPEightLo{} to \AncFloorPEightHi{} per arm (Appendix~\ref{app:grid}).
Qwen's M0 value (\RadarQwenMZeroPEight{}) rests on a zero-variance interval and a constant policy in a game where the reference is itself state-blind, so it is uninformative; its first informative reading is \RadarQwenMTwoPEight{} at M2, where its Dark Hex action varies (Fig.~\ref{fig:stateblind-all}).
Mistral's is \RadarMistralMZeroPEight{} at every level.
At M0 and M1 it is not measured for OLMo, whose Dark Hex play reached no informative node.

Appendix~\ref{app:defects} lists the measured defect rates of the stated beliefs by game.
Of the eight non-zero rates, two are the LMs' own, four come from the instrument, one comes from how the hidden state is keyed, and one stays unclassified.

The probe readings (\pa, \pd), the driven off-path readings (\pf) and the equal-value share are in Appendix~\ref{app:grid}, with the full property grid (Fig.~\ref{fig:verdicts}).

Figure~\ref{fig:radar} shows what memory changes: one panel per level on one radial scale.
The belief axes move a little: \pgb{} dips at M1 and recovers at M2 for every model, staying within \RadarOlmoMOnePSevenb{} to \RadarQwenMTwoPSevenb{}, while \pga{} closeness falls by M2 for every model (Qwen \RadarQwenMZeroPSevena{} to \RadarQwenMTwoPSevena{}, OLMo \RadarOlmoMZeroPSevena{} to \RadarOlmoMTwoPSevena{}, Mistral \RadarMistralMZeroPSevena{} to \RadarMistralMTwoPSevena{}) and does not recover.
The probe axes \pa{} and \pd{} and the calibration axis \pe{} are flat for Qwen and OLMo; Mistral's \pa{} and \pd{} fall by M2 (\RadarMistralMZeroPOne{} to \RadarMistralMTwoPOne{} and \RadarMistralMZeroPFour{} to \RadarMistralMTwoPFour{}).
What moves are the axes that depend on play against the opponent, \pc{} and \ph.
Qwen's captured value of information goes from an uninformative \RadarQwenMZeroPEight{} to \RadarQwenMTwoPEight{}, OLMo's reads only at M2, where its \RadarOlmoMTwoPEight{} is again one policy repeated, and Mistral's stays at \RadarMistralMZeroPEight{}.
The \pc{} margin moves without a direction: the cells whose mean is worse than the uniform-random player's go \RadarWorseQwenMZero{}, \RadarWorseQwenMOne{} and \RadarWorseQwenMTwo{} of \RadarCellsQwenMZero{} across the levels for Qwen, \RadarWorseOlmoMZero{}, \RadarWorseOlmoMOne{} and \RadarWorseOlmoMTwo{} for OLMo, and \RadarWorseMistralMZero{}, \RadarWorseMistralMOne{} and \RadarWorseMistralMTwo{} for Mistral.
Memory therefore changes how the agent plays against the opponent more than how its stated belief relates to its action.

\subsection{Stated beliefs are far from the equilibrium posterior, and Qwen's and Mistral's actions follow them where they respond to the state}
\label{sec:belief}
At M0 and temperature 0, stated beliefs sit far from the equilibrium posterior, \Meanv{\armQ}{PSevena} to \Meanv{\armM}{PSevena} in TVD. Where the action responds to the state, its margin over chance agreement with the stated belief is \ClRespQwen{} for Qwen, \ClRespMistral{} for Mistral on \ClRespMistralN{} cells and \ClRespOlmo{} for OLMo.

The two headline belief properties read the same node twice.
\pga{} is $\mathrm{TVD}(\hat b, \mu)$, the distance from the stated belief to the equilibrium posterior; \pgb{} asks whether the action best-responds to $\hat b$ itself.
Each is read against a floor agent scored through the same pipeline, with the Bayesian agent as ceiling (Fig.~\ref{fig:verdicts}).
At M0 the stated belief closes \AncPSevenaMZeroLo{} to \AncPSevenaMZeroHi{} of the gap from a uniform stater (\AncFloorPSevenaMZeroLo{} to \AncFloorPSevenaMZeroHi{} TVD) to the posterior.
Every arm's mean \pga{} is farther from the equilibrium posterior than a stater that never updates (\AncPriorPSevenaLo{} to \AncPriorPSevenaHi{} over the nine arms); per cell, \AncCellsPSevenaPriorAbove{} of \AncCellsPSevenaTotal{} belief-error cells beat it (Fig.~\ref{fig:verdicts}).
The action closes \AncPSevenbMZeroLo{} to \AncPSevenbMZeroHi{} of the gap from chance to full consistency.
Even the best arm's cells spread widely: its \pga{} is \PSevenAv{\armQ}{Kuhn} across the Kuhn cells and \PSevenAv{\armQ}{Goofspiel} on Goofspiel.
Its \pgb{} runs from \ClPSevenbQwenDarkhex{} on its one admitted Dark Hex cell (Table~\ref{tab:chance}; the same on either basis) and \PSevenBv{\armQ}{Liarsdice} across the Liar's Dice cells to \PSevenBv{\armQ}{Goofspiel} on Goofspiel, where state-blind play means the \pgb{} fraction measures the response format.
The belief call follows the action call and never sees it, so the order cannot condition the belief on the action; whether reversing the order changes \pgb{} is untested.
The belief prompt assumes an unexploitable opponent and the action prompt names none.
A change of framing between the two calls is therefore the main alternative explanation for low \pgb{} (Section~\ref{sec:limits}).
The three \pga{} means differ by \ClPSevenaMeanSpread{}, within the spread between cells, so \pga{} does not rank the LMs.
Agreement has a chance level that depends on the action sets: a uniformly random action lands in the best-response set at \ClChanceGameKuhn{} in Kuhn, \ClChanceGameLeduc{} in Leduc, \ClChanceGameGoofspiel{} in Goofspiel, \ClChanceGameLiarsdice{} in Liar's Dice and \ClChanceGameDarkhex{} in Dark Hex.
Over the same admissible cells the chance level is \ClChanceQwen{} for Qwen, \ClChanceOlmo{} for OLMo and \ClChanceMistral{} for Mistral (Appendix~\ref{app:chance}).
These weight the admissible cells equally; Table~\ref{tab:chance} weights by rows, so its chance levels and margins differ from these in the second decimal.
Appendix~\ref{app:optional} adds exploratory checks on \pga{}, \pgb{}, \pa{} and \pb{}.
The headline \pgb{} is on the exact basis and the margins below on the sampled basis (Appendix~\ref{app:chance} defines both).
Over the sampled nodes the margin is \ClGapSQwen{} for Qwen (95\% interval \ClGapSQwenLo{} to \ClGapSQwenHi{}), \ClGapSMistral{} for Mistral (\ClGapSMistralLo{} to \ClGapSMistralHi{}) and \ClGapSOlmo{} for OLMo (\ClGapSOlmoLo{} to \ClGapSOlmoHi{}).
The intervals are over ten seeds and understate uncertainty where play varies little between seeds.
Figure~\ref{fig:chance} (Appendix~\ref{app:chance}) shows the margin by game.
Qwen exceeds chance everywhere except Dark Hex.
Mistral exceeds it in Kuhn, Leduc, Goofspiel and Liar's Dice.
OLMo exceeds it only in Leduc, sits at it in Kuhn and falls below it in Goofspiel and Liar's Dice, so its overall margin is cancellation.
Where the action ignores the state, agreement with the best-response set is coincidence, so we also report the margin on cells whose state-blind share is below 0.5.
There the margin is \ClRespQwen{} for Qwen (\ClRespQwenN{} cells; 95\% interval \ClRespQwenLo{} to \ClRespQwenHi{}), \ClRespMistral{} for Mistral (\ClRespMistralN{} cells; \ClRespMistralLo{} to \ClRespMistralHi{}) and \ClRespOlmo{} for OLMo (\ClRespOlmoN{} cells; \ClRespOlmoLo{} to \ClRespOlmoHi{}).
This agreement is a behavioural counterpart of the belief--action gap \citet{sobotka2026broken} report for internal, probed beliefs; the two measure different things.

\pe, the per-pair calibration error minus a uniform-stating baseline (negative: better than uniform), is negative on Kuhn for every model (Qwen \PFiveMv{\armQ}{Kuhn}, OLMo \PFiveMv{\armO}{Kuhn}, Mistral \PFiveMv{\armM}{Kuhn}).
On Goofspiel one cell per model is admissible, and it is near zero (Qwen \ClPFiveGoofspielQwen{}, OLMo \ClPFiveGoofspielOlmo{}, Mistral \ClPFiveGoofspielMistral{}).
The informative cells there are state-blind on \ClSevenSBGoofspielLo{} to \ClSevenSBGoofspielHi{} of nodes.
Per cell, \pe{} lies above the uniform stater in most admissible cells of every model at M0 (Table~\ref{tab:cells}).
We do not report binned calibration error, because errors cancel within a bin.
Memory leaves the belief call intact. The belief call fails on \ClBeliefFailTotal{} nodes in all, across nine (model, memory level) arms of \ClLMNodesMin{} to \ClLMNodesMax{} LM nodes each.
The largest count is Qwen at M2 (\ClBeliefFailQwenMTwo{} of \ClLMNodesQwenMTwo{}).
\ClActionFallbackTotal{} action calls take the fallback action.

\section{Discussion}
\label{sec:limits}
At this tier, whether an LM agent's play is blind to the state depends on the game, the model and the memory condition together.
The map is a diagnostic, not a property of a model: it shows where the belief and action metrics can be interpreted, and it depends on model, prompt and decoding together.

\paragraph{What other evaluators can take from this.}
Report a chance level beside any belief--action consistency number; here it shows that OLMo's agreement does not exceed chance.
Check a result at a second decoding temperature before reading it as a property of a model.
Withhold a score where the action never varies with the state and print the reason, because such a cell describes the response format.

\paragraph{Limitations.}
With three 12B to 14B instruct LMs, one prompt format each and no reasoning model, we make no claim about language models in general.
In Goofspiel and Dark Hex the Bayesian opponent plays its fallback policy, so those games have three distinct opponents.
At temperature 0 the seeds vary only the deal, so intervals over seeds reflect deal variance and not uncertainty about the LM.
We apply no correction for the \CellsTotal{} cells and \BeliefCellsTotal{} belief scores of a collection.
Every result characterises the LM under this fixed interface; interface sensitivity is untested here.
A falling constant-action share under memory could also mean that any longer prompt knocks a greedy decoder off a fixed response.
The control, a length-matched placebo rung with the same tokens and no history, was not run.
The collapse of the tally at temperature 0.2 gives that reading weight.
The stated belief is itself a verbal report, and internal beliefs can be more accurate than what a model states~\citep{sobotka2026broken}.
At temperature 0.2 a stochastic agent can disagree with itself with no belief involved.
The belief prompt also sets out the update in three steps, so \pga{} partly measures how well the agent follows that recipe.
Because the two prompts frame the opponent differently (Section~\ref{sec:belief}), \pgb{} cannot separate a belief the action ignores from a change of framing, and a matched-prompt collection is the next stage.
The headline runs cannot tell a positional bias from a label preference; the M0 reordering probe (Appendix~\ref{app:optional}) points to label preference in most games and to neither in Liar's Dice.

\paragraph{Temperature and prompt.}
The state-blind and constant-action results describe a model, a prompt and a decoding setting together.
At temperature~0.2, with the same prompts and scorers, the belief results hold (\pga{} \ClVEightPSevenaLo{} to \ClVEightPSevenaHi{}, \pgb{} \ClVEightPSevenbLo{} to \ClVEightPSevenbHi{}, readings \ClVEightPSevenaSame{}). The state-blind results do not: the cells at 1 fall from \SBCellsCeiling{} to \ClVEightSBCeiling{} and the \ClSevenMemSup{} supported ladders become \ClEightMemSup{} (Appendix~\ref{app:temperature}).
Part of the state-blind structure therefore comes from greedy decoding, and the prompt-version-4 collection gave different shares again (Section~\ref{sec:stateblind}); the temperature was chosen after the temperature-0 results.
\paragraph{Choice of reference.}
We used CFR+ to build the instrument, not as a claim about what optimal play is: in small games it is precise, cheap and checkable, and it does not scale to deeper games.
It is one of many equilibrium references, so a distance to it ignores other equilibria, penalises actions that are equally valuable, and cannot credit deviations that exploit a non-equilibrium opponent; we read the distance as deviation from a reference, not as loss of value.
The belief properties inherit the same limit through the opponent model: \pe{} and \pga{} score the stated belief against the posterior a CFR+ opponent would induce, and \pgb{} values the best response under CFR+ continuation. Against the always-bet and RL opponents, a correct model of the opponent, or an action that exploits it, is therefore scored as a deviation.
Only \pa{}, \pb{} and \pd{}, which compare the agent with itself, are free of it, and opponents are frozen in this version.
\paragraph{Future work.}
This paper is the measurement stage of a three-stage programme: measure, explain, then intervene.
The headline run, under way and not reported here, scores instruct and reasoning models near 30B parameters within the same families over deeper games with more hands and seeds per cell.
Its pre-registration extends the anchors of Fig.~\ref{fig:verdicts}: a finer \pb{} trajectory with a hindsight adaptive agent as ceiling, the reordering probe at M1 and M2, a null band for every probe, and a tie tolerance at the solver's residual.
\ph{} needs a Dark Hex board that is both non-degenerate and exactly solvable, an open problem; the state space grows steeply with board size, and $3 \times 3$ is a known first-player win~\citep{tapkan2022darkhex}.
The mechanistic stage comes second: each logged node stores the stated belief and the equilibrium posterior as separate columns, and the probes of \citet{sobotka2026broken}, run beside our per-node scores, ask where the stated belief, the internal belief and the action diverge.
An intervention stage comes last, contingent on the mechanistic answer: property-targeted fine-tuning on verifiable rewards built from this instrument's own logs, or handing action selection to a dedicated planner, as Tru-POMDP does~\citep{tang2025trupomdp}.

\section{Conclusion}
\label{sec:conclusion}
The state-blind share of nodes varies with the game, the model and the memory condition together, and the map of that constancy is the deliverable of the measurement stage.
The stated belief closes at most \AncPSevenaMZeroHi{} of the gap from a uniform stater to the posterior and trails a stater that never updates.
Where the action responds to the state the margin over chance is \ClRespQwen{} for Qwen, \ClRespMistral{} for Mistral and \ClRespOlmo{} for OLMo.
Memory lowers the constant-action share in \ClSevenMemSup{} of the \ClSevenMemTestable{} testable ladders and raises it marginally in \ClSevenMemCon{}.
A per-cell gate withholds the belief scores a run cannot support and prints the reason for each, and every admitted score carries its defect rates.
The next stage asks whether any internal step of a language model selects an action against a belief.
\FloatBarrier
\begin{ack}
This research includes computations using the computational cluster Katana supported by Research Technology Services at UNSW Sydney (\url{https://doi.org/10.26190/669x-a286}). This research was undertaken with the assistance of resources from the National Computational Infrastructure (NCI Australia), an NCRIS enabled capability supported by the Australian Government. Michael Kingston is supported by an Australian Government Research Training Program (RTP) Scholarship.
\end{ack}

\bibliographystyle{plainnat}
\bibliography{refs}

\appendix
\section{How the five games differ}
\label{app:games}
\begin{table}[h]
\centering\small
\setlength{\tabcolsep}{2pt}
\caption{\textbf{Structural differences between the games.} Information sets count both seats over the full tree (OpenSpiel); actions is the range of legal actions over decision nodes; belief over is what the stated belief is over; moves is the longest play in one hand; words is the word count of the first-decision action and belief prompts at M0 (prompt set 7). Any of these can drive a cross-game difference, which is why cross-game differences are reported as observations. Expl.: the CFR+ reference's exploitability (NashConv halved, OpenSpiel) as a percentage of the game's largest payoff.}
\label{tab:games}
\begin{tabular}{@{}lrrrrllr@{}}
\toprule
Game & Info sets & Actions & Moves & Words (act./bel.) & Chance & Belief over (values) & Expl.\,\% \\
\midrule
Kuhn        & 12     & 2       & 3  & 232 / 422 & deal        & opponent's card (2) & \RefExploitKuhn \\
Leduc       & 936    & 2--3    & 8  & 275 / 503 & deal, public & opponent's rank (3) & \RefExploitLeduc \\
Goofspiel   & 162    & 2--4    & 6  & 215 / 408 & none        & opponent's bid (up to 4) & \RefExploitGoofspiel \\
Liar's Dice & 24,576 & 1--12   & 13 & 304 / 461 & roll        & opponent's die (6) & \RefExploitLiarsdice \\
Dark Hex    & 134    & 1--4    & 7  & 227 / 385 & none        & each cell's occupancy (4 cells) & \RefExploitDarkhex \\
\bottomrule
\end{tabular}
\end{table}

\section{Memory-ladder verdict rules}
\label{app:ladders}
Each ladder is one (model, game) pair read at M0, M1 and M2.
For each rung, the change is the rung's constant-action share minus the M0 share, so a negative change means memory lowered the share.
Two rules turn the changes into a verdict.
The population rule takes the (legal-action set, seat) bins populated in both arms, fixed once from the pooled data.
The verdict rule counts a rung only where its change exceeds the ladder's between-seed spread.
A ladder is supported when a rung clears the spread with a negative change and the ladder does not rise.
It is contrary when a rung clears the spread with a positive change.
It is indeterminate when rungs clear the spread with a negative change but the ladder rises.
It is refuted when no rung clears the spread.
Table~\ref{tab:ladders} lists all \ClSevenMemLadders{} ladders.

\begin{table}[h]
\centering\small
\setlength{\tabcolsep}{3pt}
\caption{\textbf{The memory ladders.} Constant-action share at M0, M1 and M2, the change and the between-seed spread at each rung, and the verdict (S supported, R refuted, C contrary, I indeterminate). A dagger marks a verdict that rests on a margin below 0.01; verdicts use the unrounded changes and spreads, so two rows that look equal at two decimals can resolve either way. The M0 column shows the bins shared with M1; each rung's change uses the bins shared with that rung.}
\label{tab:ladders}
\begin{tabular}{@{}llrrrrrrrc@{}}
\toprule
Model & Game & M0 & M1 & M2 & \(\Delta\)M1 & Spread M1 & \(\Delta\)M2 & Spread M2 & Verdict \\
\midrule
Qwen & Kuhn & 0.00 & 0.00 & 0.00 & 0.00 & 0.00 & 0.00 & 0.00 & R \\
Qwen & Leduc & 0.11 & 0.10 & 0.10 & $-0.01$ & 0.01 & $-0.01$ & 0.01 & R \\
Qwen & Goofspiel & 0.90 & 0.02 & 0.32 & $-0.88$ & 0.11 & $-0.59$ & 0.01 & I \\
Qwen & Liar's Dice & 0.17 & 0.08 & 0.01 & $-0.09$ & 0.11 & $-0.15$ & 0.09 & S \\
Qwen & Dark Hex & 0.90 & 0.80 & 0.00 & $-0.09$ & 0.10 & $-0.90$ & 0.10 & S \\
\midrule
OLMo & Kuhn & 1.00 & 1.00 & 0.29 & 0.00 & 0.00 & $-0.71$ & 0.00 & S \\
OLMo & Leduc & 0.74 & 0.73 & 0.73 & $-0.01$ & 0.01 & $-0.01$ & 0.19 & S\(\dagger\) \\
OLMo & Goofspiel & 0.92 & 0.75 & 0.75 & $-0.17$ & 0.17 & $-0.17$ & 0.08 & S \\
OLMo & Liar's Dice & 0.29 & 0.19 & 0.04 & $-0.10$ & 0.03 & $-0.25$ & 0.03 & S \\
OLMo & Dark Hex & 0.53 & 0.79 & 0.27 & 0.25 & 0.25 & $-0.26$ & 0.26 & C\(\dagger\) \\
\midrule
Mistral & Kuhn & 0.00 & 0.00 & 0.00 & 0.00 & 0.00 & 0.00 & 0.00 & R \\
Mistral & Leduc & 1.00 & 1.00 & 0.17 & 0.00 & 0.00 & $-0.83$ & 0.00 & S \\
Mistral & Goofspiel & 0.75 & 0.67 & 0.42 & $-0.08$ & 0.25 & $-0.33$ & 0.33 & R \\
Mistral & Liar's Dice & 0.62 & 0.66 & 0.00 & 0.04 & 0.17 & $-0.63$ & 0.22 & S \\
Mistral & Dark Hex & 1.00 & 1.00 & 0.64 & 0.00 & 0.00 & $-0.36$ & 0.18 & S \\
\bottomrule
\end{tabular}
\end{table}

\FloatBarrier
\section{Chance level and equal-value share}
\label{app:chance}
\begin{figure}[!ht]
\centering
\includegraphics[width=280pt]{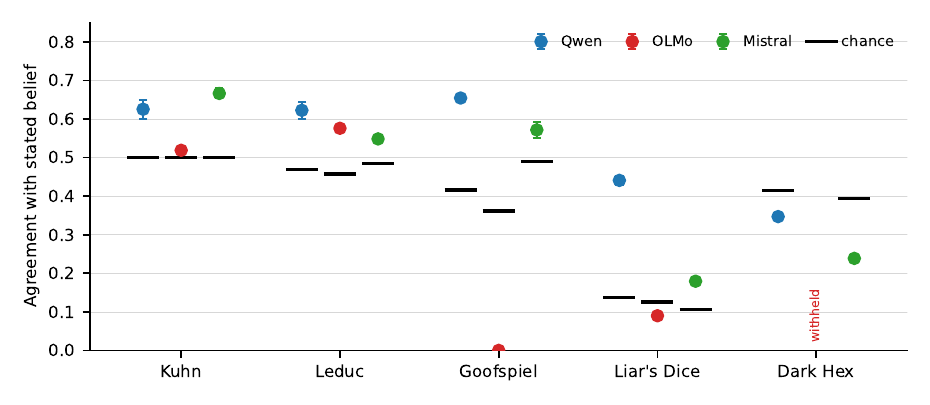}
\caption{\textbf{Qwen's action clearly follows its stated belief; Mistral's does on \ClRespMistralN{} responsive cells; OLMo's does not exceed chance.} Agreement with the stated belief per model and game, against the chance level. Points are means over ten seeds on the sampled nodes, pooled over the admissible opponents of each game; a model with no admissible cell in a game is marked withheld. Whiskers are 95\% intervals over seeds and are narrow where play is nearly deterministic; dashes mark chance.}
\label{fig:chance}
\end{figure}
The chance level of \pgb{} at a node is the share of legal actions that lie in the best-response set.
That share is the probability that a uniformly random action agrees with the stated belief, and we average it over the rows the scorer scores.
A node where every action ties agrees whatever the LM plays (All-tie).
The equal-value share is the share of LM nodes whose chosen action is within 0.01 payoff units of the best action under the CFR+ reference's value function.
Table~\ref{tab:chance} reports both for every (model, game), pooled over the admissible opponents and the ten seeds; the equal-value share pools all four opponents.
The headline \pgb{} is on the exact basis, which scores the probability mass the LM's own distribution puts on the best-response set.
The margins below use the sampled basis, which scores the logged action.
The two bases differ because the exact basis reads the probability the LM puts on each legal action in a separate log-probability pass.
At temperature 0 that distribution is not a point mass although the logged action is its mode.

\begin{table}[h]
\centering\small
\setlength{\tabcolsep}{3pt}
\caption{\textbf{\pgb{} against its chance level, and the equal-value share.} Rows pool the admissible opponents of a game, and a game with no admissible cell is withheld. \pgb{} is on the sampled basis, weighted by rows, so it differs from the headline, which averages admissible cells on the exact basis. The equal-value share pools all four opponents.}
\label{tab:chance}
\begin{tabular}{@{}llrrrrr@{}}
\toprule
Model & Game & Cells & P7b & Chance & All-tie & Equal-value \\
\midrule
Qwen & Kuhn & 4 & 0.63 & 0.50 & 0.00 & 0.78 \\
Qwen & Leduc & 4 & 0.62 & 0.47 & 0.00 & 0.88 \\
Qwen & Goofspiel & 1 & 0.65 & 0.42 & 0.11 & 0.65 \\
Qwen & Liar's Dice & 4 & 0.44 & 0.14 & 0.00 & 0.68 \\
Qwen & Dark Hex & 1 & 0.35 & 0.41 & 0.02 & 0.68 \\
\cmidrule{1-7}
Qwen & all admissible & 14 & 0.56 & 0.38 & 0.01 & 0.73 \\
\midrule
OLMo & Kuhn & 4 & 0.52 & 0.50 & 0.00 & 0.83 \\
OLMo & Leduc & 4 & 0.58 & 0.46 & 0.00 & 0.83 \\
OLMo & Goofspiel & 1 & 0.00 & 0.36 & 0.00 & 0.21 \\
OLMo & Liar's Dice & 4 & 0.09 & 0.13 & 0.00 & 0.45 \\
OLMo & Dark Hex & 0 & withheld & withheld & withheld & 0.77 \\
\cmidrule{1-7}
OLMo & all admissible & 13 & 0.35 & 0.36 & 0.00 & 0.57 \\
\midrule
Mistral & Kuhn & 4 & 0.67 & 0.50 & 0.00 & 0.80 \\
Mistral & Leduc & 4 & 0.55 & 0.48 & 0.00 & 0.88 \\
Mistral & Goofspiel & 1 & 0.57 & 0.49 & 0.27 & 0.66 \\
Mistral & Liar's Dice & 4 & 0.18 & 0.11 & 0.00 & 0.57 \\
Mistral & Dark Hex & 1 & 0.24 & 0.39 & 0.00 & 0.48 \\
\cmidrule{1-7}
Mistral & all admissible & 14 & 0.46 & 0.39 & 0.03 & 0.68 \\
\bottomrule
\end{tabular}
\end{table}

\FloatBarrier
\section{Property grid, probe readings and grounding}
\label{app:grid}
\begin{figure}[!ht]
\centering
\includegraphics[width=330pt]{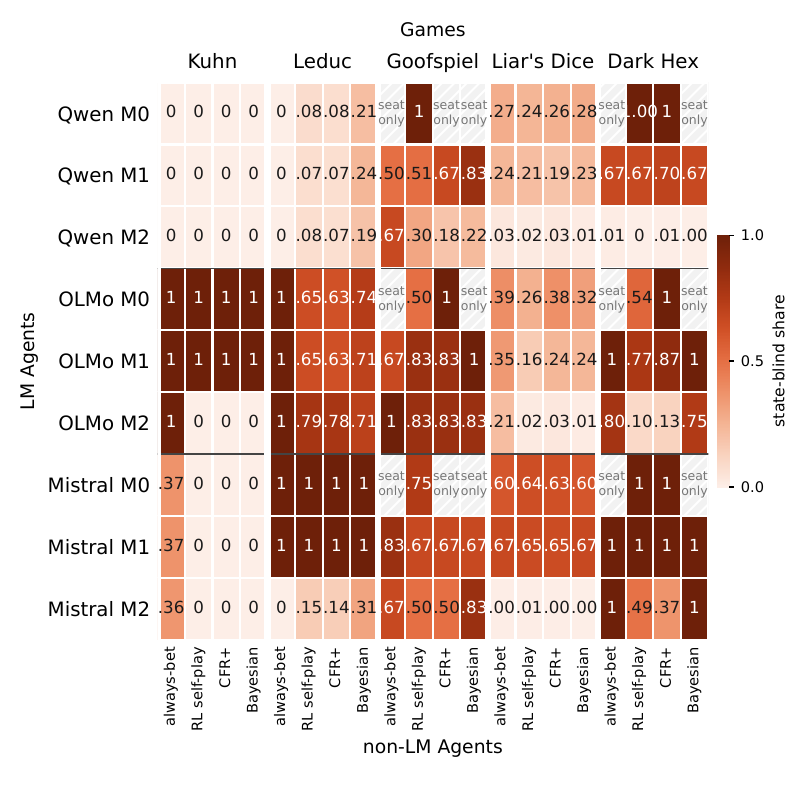}
\caption{\textbf{State-blind share at every memory level.} As Fig.~\ref{fig:stateblind}, with the M1 and M2 rows; the counts in the text use the M0 rows.}
\label{fig:stateblind-all}
\end{figure}

\begin{figure}[!ht]
\centering
\includegraphics[width=295pt]{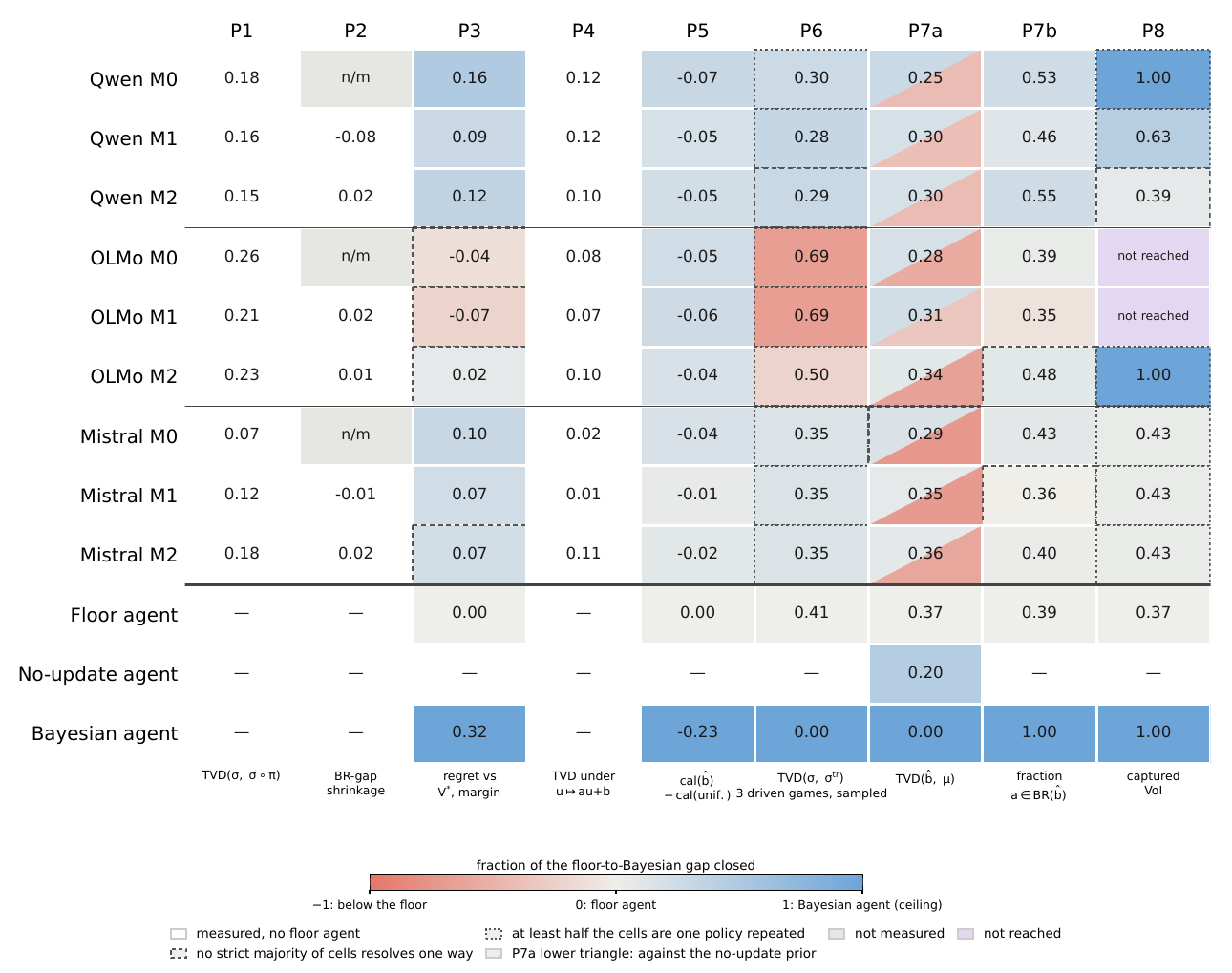}
\caption{\textbf{Of the \AncCellsTotal{} admissible cells behind the \AncColoured{} anchored tiles, \AncCellsBelow{} lie below their own floor with a seed interval that clears it, \AncCellsAbove{} above, \AncCellsStraddle{} straddle it and \AncCellsDegenerate{} are one policy repeated.} Per (model, property): the admissible-cell mean on that property's own estimand (bottom row), from the decision artifacts. Colour is the fraction of the gap from a floor agent to the Bayesian agent that the LM closes, on each arm's own admissible cells. The floors are the uniform-random player for \pc{} and, per game, for \pf{}; the uniform stater for \pe{} and \pga{}, where each \pga{} tile is split, its upper-left triangle against the uniform stater and its lower-right against the no-update prior, below which \AncCellsPSevenaPriorBelow{} of \AncCellsPSevenaTotal{} cells lie; the chance level for \pgb{}; the uniform-random capture rate for \ph{}. \pa{} under reordering is read in Appendix~\ref{app:optional} and not here: its ceiling is reached by uniform-random, constant-name and Bayesian play alike, the reason \pa{} under relabelling carries no floor agent. The three bottom rows print the floors, the no-update prior and the ceilings pooled over the nine arms; each arm's colour uses its own floors, so a cell's colour can differ from what the printed rows imply. Resolution is counted per cell against the cell's own floor and its Student-$t$ interval over seeds, never from the arm-level envelope, which spans the cells. A tile's edge summarises its cells by a strict majority of all of them. Solid: more than half resolve on one side. Dashed: no side has more than half (\AncUnresolvedN{} tiles: \AncUnresolvedTiles{}). Dotted: at least half are one policy repeated. So \AncBelowResolved{} tiles resolve below their floor (\AncBelowTiles{}) and \AncAboveResolved{} above; Table~\ref{tab:cells} prints the counts. Colour is the mean anchored score and the count is per cell, so the two can disagree: \AncConflictTiles{} read blue while most of their cells sit just below the floor. The resolved intervals are narrow (median width \AncResolvedWidthMedian{} over ten seeds), so resolution is a sign test at this precision. The median distance from the floor is \AncBelowGapMedian{} for the cells below and \AncAboveGapMedian{} for those above; \AncBelowWithinTwo{} of the \AncCellsBelow{} below cells lie within 0.02 of it. \pa{} under relabelling, \pd{} and \pb{} stay white: a uniform-random agent is invariant under renaming and rescaling by construction, and the adaptive agent of Appendix~\ref{app:optional} adapts inside \pb{}'s first bin, so its two-bin delta sets no ceiling above the floor. The null band of \pa{} under relabelling and \pd{}, the same paired distance between two base-arm seeds, is \ClNullBandQwen{} for Qwen, \ClNullBandOlmo{} for OLMo and \ClNullBandMistral{} for Mistral; no game pair exceeds \ClNullBandMaxPair{}. \pgb{} is on the exact basis at M0 and the sampled basis at M1 and M2; the chance level is the same on both. On the $2 \times 2$ board the CFR+ reference reaches the informative \ph{} nodes with negligible mass and plays uniformly there, so as played it captures at the floor rate (\AncAllPEightRef{}); the ceiling is the best response to the equilibrium posterior. \pf{} is the mean over the Kuhn, Leduc and Liar's Dice driven cells on the sampled basis, each against its own game's random player; the \pf{} paragraph below reads the exact basis at M0, where OLMo's Kuhn cell clears the random player, and a greedy point mass cannot match a mixed reference under TVD, so part of any \pf{} distance comes from playing a single action. The pooling crosses games, so a tile can read above the floor while one game fails (Mistral on Kuhn). Counted per driven game, every \pf{} tile but Qwen at M2 has at least half its games one policy repeated. Qwen's one-policy games are Leduc and Liar's Dice at M0 and M1 and Liar's Dice at M2. OLMo's one-policy games are all three at M0 and M1 and Kuhn and Leduc at M2. Mistral's one-policy games are all three at every level. Qwen at M2 has no majority either way, so no \pf{} tile resolves. At game level on the sampled basis, \PSixGameBelowN{} driven cell with a non-degenerate interval lies below its own random player (\PSixGameBelow{}) and \PSixGameAboveN{} above (\PSixGameAbove{}); on the exact basis at M0, OLMo's Leduc and Liar's Dice cells and Mistral's Kuhn cell lie below theirs (below). Every \pc{} and \pf{} reading is a distance from the reference's mixture, not value lost (Section~\ref{sec:limits}). Qwen's M0 \ph{} value of \RadarQwenMZeroPEight{} rests on a constant action that happens to capture (Section~\ref{sec:admiss}), not on Bayesian play.}
\label{fig:verdicts}
\end{figure}

Figure~\ref{fig:stateblind-all} repeats the state-blind grid of Fig.~\ref{fig:stateblind} with the M1 and M2 rows (2000 hands $\times$ 10 seeds), and Fig.~\ref{fig:verdicts} prints every (model, property) value at M0, M1 and M2, coloured by the fraction of the gap from a floor agent to the Bayesian agent that the LM closes; Table~\ref{tab:anchors} lists the fractions.
\begin{table}[!ht]
\centering\footnotesize
\setlength{\tabcolsep}{5pt}
\caption{\textbf{Anchored scores behind Fig.~\ref{fig:verdicts}.} Fraction of the gap from the floor agent to the Bayesian agent that the LM closes, per arm and property, on each arm's own admissible cells; negative is below the floor. A dagger marks a tile with no strict majority of its cells on either side of the floor. A double dagger marks one at least half of whose cells are one policy repeated. The \pga{} (prior) column anchors against the no-update prior. The bottom rows are the pooled floors, the no-update prior and the ceilings, on each property's own estimand.}
\label{tab:anchors}
\begin{tabular}{@{}llrrrrrrr@{}}
\toprule
Model & Level & \pc{} & \pe{} & \pf{} & \pga{} & \pga{} (prior) & \pgb{} & \ph{} \\
\midrule
Qwen & M0 & 0.50 & 0.31 & 0.27$^{\ddagger}$ & 0.29 & $-$0.43 & 0.24 & 1.00$^{\ddagger}$ \\
Qwen & M1 & 0.29 & 0.19 & 0.32$^{\ddagger}$ & 0.19 & $-$0.40 & 0.12 & 0.43$^{\ddagger}$ \\
Qwen & M2 & 0.36 & 0.22 & 0.30$^{\dagger}$ & 0.19 & $-$0.42 & 0.25 & 0.08$^{\dagger}$ \\
\midrule
OLMo & M0 & $-$0.14$^{\dagger}$ & 0.24 & $-$0.68$^{\ddagger}$ & 0.19 & $-$0.57 & 0.05 & n/m \\
OLMo & M1 & $-$0.23$^{\dagger}$ & 0.25 & $-$0.68$^{\ddagger}$ & 0.22 & $-$0.34 & $-$0.10 & n/m \\
OLMo & M2 & 0.08$^{\dagger}$ & 0.17 & $-$0.23$^{\ddagger}$ & 0.10 & $-$0.65 & 0.10$^{\dagger}$ & 1.00$^{\ddagger}$ \\
\midrule
Mistral & M0 & 0.31 & 0.18 & 0.15$^{\ddagger}$ & 0.16$^{\dagger}$ & $-$0.73$^{\dagger}$ & 0.09 & 0.05$^{\ddagger}$ \\
Mistral & M1 & 0.23 & 0.06 & 0.15$^{\ddagger}$ & 0.07 & $-$0.70 & $-$0.00$^{\dagger}$ & 0.04$^{\ddagger}$ \\
Mistral & M2 & 0.22$^{\dagger}$ & 0.10 & 0.15$^{\ddagger}$ & 0.08 & $-$0.60 & 0.03 & 0.04$^{\ddagger}$ \\
\midrule
Floor agent & & 0.00 & 0.00 & 0.41 & 0.37 & 0.20 & 0.39 & 0.37 \\
No-update agent & & n/a & n/a & n/a & 0.20 & 0.20 & n/a & n/a \\
Bayesian agent & & 0.32 & $-$0.23 & 0.00 & 0.00 & 0.00 & 1.00 & 1.00 \\
\bottomrule
\end{tabular}
\end{table}
\begin{table}[!ht]
\centering\footnotesize
\setlength{\tabcolsep}{5pt}
\caption{\textbf{The cells behind each tile of Fig.~\ref{fig:verdicts}.} Per arm and property: admissible cells below their own floor with a seed interval that clears it / above it / straddling it / one policy repeated. \pf{} counts driven games. One policy repeated means the LM cell, whatever the floor's interval; by property these are \pc{} \AncCellsPThreeDegenerate{}, \pf{} \AncCellsPSixDegenerate{}, \pgb{} \AncCellsPSevenbDegenerate{} and \ph{} \AncCellsPEightDegenerate{}. A tile resolves when more than half of all its cells lie on one side of the floor.}
\label{tab:cells}
\begin{tabular}{@{}llrrrrrr@{}}
\toprule
Model & Level & \pc{} & \pe{} & \pf{} & \pga{} & \pgb{} & \ph{} \\
\midrule
Qwen & M0 & 0/16/0/4 & 1/13/0/0 & 0/0/1/2 & 1/13/0/0 & 1/13/0/0 & 0/0/0/4 \\
Qwen & M1 & 6/12/0/2 & 6/14/0/0 & 0/0/1/2 & 6/14/0/0 & 7/13/0/0 & 0/2/0/2 \\
Qwen & M2 & 4/15/1/0 & 2/18/0/0 & 0/1/1/1 & 2/17/1/0 & 1/19/0/0 & 2/2/0/0 \\
\midrule
OLMo & M0 & 7/9/2/2 & 4/9/0/0 & 0/0/0/3 & 4/9/0/0 & 4/9/0/0 & n/m \\
OLMo & M1 & 9/7/1/3 & 4/15/0/0 & 0/0/0/3 & 4/15/0/0 & 10/7/0/2 & n/m \\
OLMo & M2 & 4/6/6/4 & 9/10/0/0 & 1/0/0/2 & 9/10/0/0 & 9/9/0/1 & 0/0/0/2 \\
\midrule
Mistral & M0 & 2/12/3/3 & 5/8/1/0 & 0/0/0/3 & 5/7/2/0 & 2/12/0/0 & 1/1/0/2 \\
Mistral & M1 & 3/13/2/2 & 11/8/1/0 & 0/0/0/3 & 11/8/1/0 & 7/9/2/2 & 1/1/0/2 \\
Mistral & M2 & 4/10/2/4 & 7/11/2/0 & 0/0/0/3 & 8/11/1/0 & 5/15/0/0 & 1/1/0/2 \\
\bottomrule
\end{tabular}
\end{table}

Relabelling (\pa) and payoff scaling (\pd) come from the paired probe runs and have no floor agent.
Their null band, the same paired distance between two base-arm seeds, is measured at M0 per game and reaches \ClNullBandQwenKuhnMax{} in one Kuhn seed pair.
Every non-zero reading below exceeds its game's band except OLMo's \pd{} in Dark Hex (\ClPFourOlmoDarkhex{} against a band of up to \ClNullBandOlmoDarkhexMax{}).
Mistral's band is zero, so its readings clear it by construction.
Under option reordering \pa{} does have a floor agent (Appendix~\ref{app:optional}).
Relabelling shifts the policy by at most \POnev{\armQ}{Liarsdice} unweighted paired TVD on Qwen, \POnev{\armO}{Leduc} on OLMo and \POnev{\armM}{Liarsdice} on Mistral outside Dark Hex.
In Dark Hex a partial relabelling moves Qwen by \POnev{\armQ}{Darkhex}, OLMo by \POnev{\armO}{Darkhex} and Mistral by \POnev{\armM}{Darkhex}.
On Kuhn, payoff scaling does not move OLMo or Mistral (\PFourv{\armO}{Kuhn}, \PFourv{\armM}{Kuhn}) and moves Qwen by \PFourv{\armQ}{Kuhn}; over every measurable game the means are \PFourheadv{\armO}, \PFourheadv{\armM} and \PFourheadv{\armQ} (Fig.~\ref{fig:verdicts}).

Sequential rationality (\pf) is read at driven nodes instead, on the exact distribution basis against a uniform-random baseline for each game (Kuhn \ClPSixBaseKuhn{}, Leduc \ClPSixBaseLeduc{}, Liar's Dice \ClPSixBaseLiarsdice{}).
It lies below the random player, with its interval clearing the random player's, for OLMo on Kuhn (\ClPSixOlmoKuhn{}) and for Qwen (\ClPSixQwenLiarsdice{}) and Mistral (\ClPSixMistralLiarsdice{}) on Liar's Dice.
It lies above it, the same way, for Mistral on Kuhn (\ClPSixMistralKuhn{}) and for OLMo on Leduc (\ClPSixOlmoLeduc{}) and Liar's Dice (\ClPSixOlmoLiarsdice{}).
It is indeterminate in the other three cells.
Goofspiel and Dark Hex are withheld, because the off-path policy is uniform at every scored node.
Fig.~\ref{fig:verdicts} reads the sampled basis over the other three games at every memory level.
The random player's distance differs by basis (Liar's Dice \ClPSixBaseLiarsdice{} exact against \PSixSentinelLiarsdice{} sampled) because the sampled basis scores one logged action per node and the exact basis the whole distribution.
The distance is from the CFR+ reference's mixture and does not measure the value lost against the opponent.
Judged by value, the chosen action is within 0.01 of the best action under the CFR+ reference at \ClEqualValueKuhn{} of Kuhn nodes and \ClEqualValueLeduc{} of Leduc nodes.
The shares are \ClEqualValueGoofspiel{} in Goofspiel, \ClEqualValueLiarsdice{} in Liar's Dice and \ClEqualValueDarkhex{} in Dark Hex.
A distance from the CFR+ policy therefore overstates the loss of value (Appendix~\ref{app:chance}).

The result behind each property: invariance (\pa) and payoff sensitivity (\pd) to description and to affine payoff maps~\citep{vonneumann1944theory,tversky1981framing}; adaptivity (\pb) to rational learning~\citep{kalai1993rational}; randomisation (\pc) to the mixed equilibrium~\citep{vonneumann1928theorie}; calibration (\pe) to calibrated forecasting~\citep{brier1950verification,dawid1982wellcalibrated,foster1998asymptotic}; sequential rationality (\pf) to trembling-hand and sequential equilibrium~\citep{selten1975reexamination,kreps1982sequential}; belief error and belief--action consistency (\pga, \pgb) to Bayesian games~\citep{harsanyi1967games}; value of information (\ph) to decision analysis~\citep{howard1966information}.

\begin{table}[!ht]
\centering\footnotesize
\setlength{\tabcolsep}{3pt}
\caption{\textbf{LM and reference state-blind share per informative cell at M0.} The within-seat state-blind share of the LM (LM) and of the CFR+ reference on the same groups (Ref.), rounded to two decimals, and their difference (LM minus Ref.). A positive difference is constancy the reference does not share; a cell with Ref.\ at 1.00 is reference-blind and is excluded from the \SBCellsRefVaries{}-cell counts in the text. Seat-only cells are omitted. The reference's action at a node is its most likely action under the CFR+ policy, with ties going to the lowest-numbered action; on the full mixed policy, one further cell (OLMo against RL self-play in Goofspiel) has a reference that varies.}
\label{tab:refbaseline}
\begin{tabular}{@{}lllrrr@{\hspace{10pt}}lllrrr@{}}
\toprule
Model & Game & Opponent & LM & Ref. & Diff. & Model & Game & Opponent & LM & Ref. & Diff. \\
\midrule
Qwen & Kuhn & always-bet & 0.00 & 0.00 & 0.00 & OLMo & Goofspiel & RL self-play & 0.50 & 1.00 & $-$0.50 \\
Qwen & Kuhn & RL self-play & 0.00 & 0.00 & 0.00 & OLMo & Goofspiel & CFR+ & 1.00 & 1.00 & 0.00 \\
Qwen & Kuhn & CFR+ & 0.00 & 0.00 & 0.00 & OLMo & Liar's Dice & always-bet & 0.39 & 0.04 & 0.35 \\
Qwen & Kuhn & Bayesian & 0.00 & 0.00 & 0.00 & OLMo & Liar's Dice & RL self-play & 0.26 & 0.03 & 0.23 \\
Qwen & Leduc & always-bet & 0.00 & 0.00 & 0.00 & OLMo & Liar's Dice & CFR+ & 0.38 & 0.04 & 0.34 \\
Qwen & Leduc & RL self-play & 0.08 & 0.00 & 0.08 & OLMo & Liar's Dice & Bayesian & 0.32 & 0.02 & 0.30 \\
Qwen & Leduc & CFR+ & 0.08 & 0.00 & 0.08 & OLMo & Dark Hex & RL self-play & 0.54 & 1.00 & $-$0.46 \\
Qwen & Leduc & Bayesian & 0.21 & 0.00 & 0.21 & OLMo & Dark Hex & CFR+ & 1.00 & 1.00 & 0.00 \\
Qwen & Goofspiel & RL self-play & 1.00 & 0.00 & 1.00 & Mistral & Kuhn & always-bet & 0.37 & 0.00 & 0.37 \\
Qwen & Liar's Dice & always-bet & 0.27 & 0.15 & 0.12 & Mistral & Kuhn & RL self-play & 0.00 & 0.00 & 0.00 \\
Qwen & Liar's Dice & RL self-play & 0.24 & 0.16 & 0.08 & Mistral & Kuhn & CFR+ & 0.00 & 0.00 & 0.00 \\
Qwen & Liar's Dice & CFR+ & 0.26 & 0.19 & 0.07 & Mistral & Kuhn & Bayesian & 0.00 & 0.00 & 0.00 \\
Qwen & Liar's Dice & Bayesian & 0.28 & 0.22 & 0.06 & Mistral & Leduc & always-bet & 1.00 & 0.00 & 1.00 \\
Qwen & Dark Hex & RL self-play & 1.00 & 1.00 & 0.00 & Mistral & Leduc & RL self-play & 1.00 & 0.00 & 1.00 \\
Qwen & Dark Hex & CFR+ & 1.00 & 1.00 & 0.00 & Mistral & Leduc & CFR+ & 1.00 & 0.00 & 1.00 \\
OLMo & Kuhn & always-bet & 1.00 & 0.00 & 1.00 & Mistral & Leduc & Bayesian & 1.00 & 0.00 & 1.00 \\
OLMo & Kuhn & RL self-play & 1.00 & 0.00 & 1.00 & Mistral & Goofspiel & RL self-play & 0.75 & 0.00 & 0.75 \\
OLMo & Kuhn & CFR+ & 1.00 & 0.00 & 1.00 & Mistral & Liar's Dice & always-bet & 0.60 & 0.00 & 0.60 \\
OLMo & Kuhn & Bayesian & 1.00 & 0.00 & 1.00 & Mistral & Liar's Dice & RL self-play & 0.64 & 0.01 & 0.63 \\
OLMo & Leduc & always-bet & 1.00 & 0.00 & 1.00 & Mistral & Liar's Dice & CFR+ & 0.63 & 0.03 & 0.60 \\
OLMo & Leduc & RL self-play & 0.65 & 0.00 & 0.65 & Mistral & Liar's Dice & Bayesian & 0.60 & 0.00 & 0.60 \\
OLMo & Leduc & CFR+ & 0.63 & 0.00 & 0.63 & Mistral & Dark Hex & RL self-play & 1.00 & 1.00 & 0.00 \\
OLMo & Leduc & Bayesian & 0.74 & 0.00 & 0.74 & Mistral & Dark Hex & CFR+ & 1.00 & 1.00 & 0.00 \\
\bottomrule
\end{tabular}
\end{table}

\FloatBarrier
\section{Robustness to temperature}
\label{app:temperature}
The second collection reuses the prompts, scorers and games of the first and differs only in temperature, 0.2 against 0.
Table~\ref{tab:epochs} compares the two.
The belief measures \pga{} and \pgb{} move little and keep their readings.
The state-blind counts, OLMo's state-blind shares and the memory tally move a lot.

\begin{table}[h]
\centering\small
\setlength{\tabcolsep}{3pt}
\caption{\textbf{The temperature-0 and temperature-0.2 collections.} Both use prompt set 7, ten seeds per arm and the same scorers; \pga{} and \pgb{} are the arm-level means over admissible cells.}
\label{tab:epochs}
\begin{tabular}{@{}lrr@{}}
\toprule
Quantity & Temperature 0 & Temperature 0.2 \\
\midrule
P7a, Qwen & 0.250 & 0.252 \\
P7a, OLMo & 0.283 & 0.298 \\
P7a, Mistral & 0.288 & 0.284 \\
P7b, Qwen & 0.528 & 0.535 \\
P7b, OLMo & 0.390 & 0.434 \\
P7b, Mistral & 0.431 & 0.402 \\
State-blind cells assessed & 46 & 50 \\
of them at 1 & 16 & 5 \\
of them at 0 & 8 & 21 \\
of them in between & 22 & 24 \\
Seat-only cells & 14 & 10 \\
State-blind share, OLMo on Kuhn & 1.00 & 0.00 \\
State-blind share, OLMo on Leduc & 0.63--1.00 & 0.00--0.01 \\
Belief scores admitted & 123 of 180 & 147 of 180 \\
Withheld belief cells per property (structural / thin / failed) & 12 / 2 / 5 & 8 / 2 / 1 \\
Memory ladders S / R / C / I & 9 / 4 / 1 / 1 & 4 / 10 / 0 / 1 \\
\bottomrule
\end{tabular}
\end{table}

\FloatBarrier
\section{Defect rates of the stated beliefs}
\label{app:defects}
Beside each belief score the gate prints five measured defect rates.
Table~\ref{tab:blockers} gives the non-zero rates per game as ranges over opponents; null-belief, zero-posterior and aligned key-set rates are zero throughout.
A rate is \emph{instrument-side} when the scorer or the rule that separates trivial from tracking beliefs produces it.
It is \emph{representational} when the belief encoding produces it identically across LMs.
It is \emph{model-side} when it varies across LMs under one encoding.
Of the eight rates, two are the LMs' own.
Four are instrument-side.
On Kuhn, the identity rate of Mistral and Qwen is a first-node restatement of the trivial prior, which is the correct belief there; OLMo almost never states it exactly.
Leduc's raw key-set rate fires because the scorers compare three ranks instead of six card keys.
The Leduc and Goofspiel identity rates restate the card-conditional prior at the first node.
The rule, written for a uniform prior, counts that as tracking, so the gap lies in the rule.
One rate is representational: Goofspiel's raw key-set rate, identical across LMs, reflects how the hidden state is keyed.
Two are model-side. Liar's Dice's support rate differs across the LMs under one belief representation. Leduc's support rate comes almost entirely from round-2 nodes where the LM's card pairs the public card, so the opponent cannot hold that rank. The belief prompt states that no card of that rank remains unseen and that such a rank gets 0, yet the LMs give it mass, at rates that differ across the models. The prompt's format example shows three non-zero values, which may encourage this but does not override the stated rule.
The Goofspiel support rate stays unclassified. In Goofspiel the LMs state beliefs over the full deck while the equilibrium posterior omits cards the opponent cannot hold, and the scorer counts that mass as belief error by design. We did not check whether the prompt lets a model deduce those cards, so the rate is not attributed to the LMs.

\begin{table}[!ht]
\centering\small
\setlength{\tabcolsep}{4pt}
\caption{\textbf{Defect rates of the stated beliefs per model and game} (share of belief rows, 1000 hands $\times$ 10 seeds per cell, range over four opponents, over every gated cell whether admitted or withheld; -- is zero). Identity: belief byte-equal to the posterior. Support: belief and posterior put positive mass on different sets. Raw key-set: key sets compared before the scorers' alignment. Tags: [i] instrument-side, [r] representational, [m] model-side, [u] unclassified, [t] first-node restatement of a non-uniform prior (counted as tracking by the rule).}
\label{tab:blockers}
\begin{tabular}{@{}llrrr@{}}
\toprule
Game & Defect Rate & Mistral & OLMo & Qwen \\
\midrule
Kuhn & Identity [i] & 37--43\% & 0.1--0.2\% & 43--46\% \\
Leduc & Identity [t] & 11--16\% & 14--16\% & 16--25\% \\
 & Raw Key-Set [i] & 100\% & 100\% & 100\% \\
 & Support [m] & 7--10\% & 7--10\% & 5--8\% \\
Goofspiel & Identity [t] & -- & -- & <0.1\% \\
 & Raw Key-Set [r] & 33--67\% & 33--67\% & 33--67\% \\
 & Support [u] & 67--67\% & 77--81\% & 44--56\% \\
Liar's Dice & Support [m] & 6--10\% & 2--4\% & 1--3\% \\
\bottomrule
\end{tabular}
{}
\end{table}

\FloatBarrier
\section{Exploratory analyses added after review}
\label{app:optional}
Seven analyses run after the reviews; they came after the fix list that drove this revision and are exploratory.
The first five pool over the admissible \pgb{} cells of Table~\ref{tab:chance} and reuse the scorer of Section~\ref{sec:setup} unchanged; the sixth reads the option-reordering probe, and the seventh runs an adaptive agent through the same pipeline.

\paragraph{Baselines for belief error.}
Table~\ref{tab:p7abase} gives \pga{} a scale by scoring two stand-in beliefs on the same rows as the stated belief: a uniform belief over the keys the LM was asked to state, and the no-update prior, the equilibrium posterior at the seat's first node of the hand carried forward unchanged.
Cells are weighted equally, as in the headline, so the stated column reproduces Section~\ref{sec:belief}.
The uniform stater sits \OptBaseUniformMistralAll{} to \OptBaseUniformQwenAll{} from the equilibrium posterior, and the no-update prior \OptBasePriorMistralAll{} to \OptBasePriorOlmoAll{}.
The stated belief beats the uniform stater in \OptBaseBeatsUniform{} of the \OptBasePairs{} measurable (model, game) pairs and ties or trails it in the other \OptBaseNotBeatsUniform{} (\OptBaseNotBeatsUniformList{}), and it trails the no-update prior in every pair except \OptBaseBeatsPriorList{}.
The gap is widest in Kuhn, where the prior sits at \OptBasePriorQwenKuhn{} to \OptBasePriorOlmoKuhn{} and the stated belief at \OptBaseStatedQwenKuhn{} to \OptBaseStatedOlmoKuhn{}.
An agent that stated the deal prior and never updated would therefore score better on \pga{} than any of the three LMs.
In Dark Hex the prior is the empty board, which is why it scores low there.

\begin{table}[!ht]
\centering\small
\setlength{\tabcolsep}{4pt}
\caption{\textbf{\pga{} against two baselines.} TVD from the equilibrium posterior of the stated belief, of a uniform belief over the stated keys, and of the no-update prior, per (model, game) over the admissible cells, cells weighted equally; the all-admissible rows are the headline \pga{} and its baselines.}
\label{tab:p7abase}
\begin{tabular}{@{}llrrrr@{}}
\toprule
Model & Game & Cells & Stated & Uniform & No-update prior \\
\midrule
Qwen & Kuhn & 4 & 0.18 & 0.43 & 0.05 \\
Qwen & Leduc & 4 & 0.22 & 0.30 & 0.20 \\
Qwen & Goofspiel & 1 & 0.65 & 0.67 & 0.63 \\
Qwen & Liar's Dice & 4 & 0.23 & 0.27 & 0.18 \\
Qwen & Dark Hex & 1 & 0.34 & 0.31 & 0.11 \\
\cmidrule{1-6}
Qwen & all admissible & 14 & 0.250 & 0.353 & 0.175 \\
\midrule
OLMo & Kuhn & 4 & 0.23 & 0.43 & 0.08 \\
OLMo & Leduc & 4 & 0.26 & 0.29 & 0.15 \\
OLMo & Goofspiel & 1 & 0.46 & 0.60 & 0.67 \\
OLMo & Liar's Dice & 4 & 0.31 & 0.26 & 0.19 \\
OLMo & Dark Hex & 0 & withheld & withheld & withheld \\
\cmidrule{1-6}
OLMo & all admissible & 13 & 0.283 & 0.349 & 0.180 \\
\midrule
Mistral & Kuhn & 4 & 0.23 & 0.43 & 0.07 \\
Mistral & Leduc & 4 & 0.31 & 0.30 & 0.21 \\
Mistral & Goofspiel & 1 & 0.66 & 0.67 & 0.63 \\
Mistral & Liar's Dice & 4 & 0.22 & 0.21 & 0.13 \\
Mistral & Dark Hex & 1 & 0.36 & 0.35 & 0.08 \\
\cmidrule{1-6}
Mistral & all admissible & 14 & 0.288 & 0.342 & 0.167 \\
\bottomrule
\end{tabular}
\end{table}

\paragraph{Tie tolerance for belief--action consistency.}
\pgb{} counts an action as a best response when its value under the stated belief is within $10^{-6}$ of the best, while the equal-value share of Appendix~\ref{app:chance} uses 0.01 payoff units.
Table~\ref{tab:p7btol} re-scores \pgb{} and its chance level with ties at 0.01.
Both rise, because the best-response set grows, and the margin over chance moves little: \OptTolFineMarginQwen{} to \OptTolCoarseMarginQwen{} for Qwen, \OptTolFineMarginOlmo{} to \OptTolCoarseMarginOlmo{} for OLMo and \OptTolFineMarginMistral{} to \OptTolCoarseMarginMistral{} for Mistral, where Liar's Dice accounts for the change.
Near-ties scored as misses therefore do not explain the low \pgb{}; the readings of Section~\ref{sec:belief} stand at either tolerance.

\begin{table}[!ht]
\centering\small
\setlength{\tabcolsep}{4pt}
\caption{\textbf{\pgb{} at two tie tolerances.} Agreement with the best-response set, its chance level and the margin, per (model, game) over the admissible cells weighted by rows, with ties judged at the scorer's $10^{-6}$ and at 0.01 payoff units.}
\label{tab:p7btol}
\begin{tabular}{@{}llrrrrrrr@{}}
\toprule
Model & Game & Cells & P7b & Chance & Margin & P7b & Chance & Margin \\
 & & & \multicolumn{3}{c}{tolerance $10^{-6}$} & \multicolumn{3}{c}{tolerance 0.01} \\
\midrule
Qwen & Kuhn & 4 & 0.63 & 0.50 & 0.13 & 0.78 & 0.65 & 0.13 \\
Qwen & Leduc & 4 & 0.62 & 0.47 & 0.15 & 0.63 & 0.58 & 0.05 \\
Qwen & Goofspiel & 1 & 0.65 & 0.42 & 0.24 & 0.65 & 0.42 & 0.24 \\
Qwen & Liar's Dice & 4 & 0.44 & 0.14 & 0.30 & 0.51 & 0.19 & 0.32 \\
Qwen & Dark Hex & 1 & 0.35 & 0.41 & $-$0.07 & 0.35 & 0.41 & $-$0.07 \\
\cmidrule{1-9}
Qwen & all admissible & 14 & 0.56 & 0.38 & 0.18 & 0.61 & 0.47 & 0.15 \\
\midrule
OLMo & Kuhn & 4 & 0.52 & 0.50 & 0.02 & 0.52 & 0.50 & 0.02 \\
OLMo & Leduc & 4 & 0.58 & 0.46 & 0.12 & 0.66 & 0.56 & 0.11 \\
OLMo & Goofspiel & 1 & 0.00 & 0.36 & $-$0.36 & 0.00 & 0.36 & $-$0.36 \\
OLMo & Liar's Dice & 4 & 0.09 & 0.13 & $-$0.04 & 0.13 & 0.15 & $-$0.02 \\
OLMo & Dark Hex & 0 & withheld & & & withheld & & \\
\cmidrule{1-9}
OLMo & all admissible & 13 & 0.35 & 0.36 & $-$0.01 & 0.39 & 0.40 & $-$0.01 \\
\midrule
Mistral & Kuhn & 4 & 0.67 & 0.50 & 0.17 & 0.80 & 0.63 & 0.17 \\
Mistral & Leduc & 4 & 0.55 & 0.48 & 0.06 & 0.62 & 0.56 & 0.06 \\
Mistral & Goofspiel & 1 & 0.57 & 0.49 & 0.08 & 0.57 & 0.49 & 0.08 \\
Mistral & Liar's Dice & 4 & 0.18 & 0.11 & 0.07 & 0.43 & 0.22 & 0.21 \\
Mistral & Dark Hex & 1 & 0.24 & 0.39 & $-$0.16 & 0.24 & 0.39 & $-$0.16 \\
\cmidrule{1-9}
Mistral & all admissible & 14 & 0.46 & 0.39 & 0.07 & 0.58 & 0.47 & 0.10 \\
\bottomrule
\end{tabular}
\end{table}

\paragraph{Belief error on updated nodes.}
Table~\ref{tab:p7aupd} repeats the baselines on the rows after each seat's first decision node of the hand, where the equilibrium posterior has moved on at least one opponent action.
There the stated belief sits \OptUpdStatedQwenAll{} to \OptUpdStatedMistralAll{} from the posterior, against \OptBaseStatedQwenAll{} to \OptBaseStatedMistralAll{} over all nodes, and the no-update prior sits \OptUpdPriorMistralAll{} to \OptUpdPriorQwenAll{}.
On updated nodes the stated belief is therefore about as far from the posterior as a belief that never updates, and for Qwen in Kuhn it is that belief (\OptUpdStatedQwenKuhn{} against \OptUpdPriorQwenKuhn{}).
The headline \pga{} is held down by first-node restatements of the prior, which the scorer counts in full.

\begin{table}[!ht]
\centering\small
\setlength{\tabcolsep}{4pt}
\caption{\textbf{\pga{} on updated nodes only.} TVD from the equilibrium posterior on the rows after each seat's first node of the hand, for the stated belief, a uniform belief and the no-update prior, per (model, game) over the admissible cells weighted equally; the stated value over all nodes is repeated for comparison.}
\label{tab:p7aupd}
\begin{tabular}{@{}llrrrrrr@{}}
\toprule
Model & Game & Cells & Rows & Stated, all & Stated & Uniform & No-update prior \\
 & & & updated & nodes & \multicolumn{3}{c}{updated nodes only} \\
\midrule
Qwen & Kuhn & 4 & 4,167 & 0.18 & 0.50 & 0.67 & 0.50 \\
Qwen & Leduc & 4 & 54,638 & 0.22 & 0.29 & 0.35 & 0.36 \\
Qwen & Goofspiel & 1 & 20,000 & 0.65 & 0.62 & 0.62 & 0.95 \\
Qwen & Liar's Dice & 4 & 20,448 & 0.23 & 0.44 & 0.56 & 0.54 \\
Qwen & Dark Hex & 1 & 5,311 & 0.34 & 0.34 & 0.19 & 0.31 \\
\cmidrule{1-8}
Qwen & all admissible & 14 & 104,564 & 0.250 & 0.420 & 0.511 & 0.491 \\
\midrule
OLMo & Kuhn & 4 & 11,132 & 0.23 & 0.42 & 0.55 & 0.38 \\
OLMo & Leduc & 4 & 46,353 & 0.26 & 0.36 & 0.35 & 0.27 \\
OLMo & Goofspiel & 1 & 20,000 & 0.46 & 0.55 & 0.52 & 1.00 \\
OLMo & Liar's Dice & 4 & 30,614 & 0.31 & 0.46 & 0.46 & 0.46 \\
OLMo & Dark Hex & 0 & 0 & withheld & withheld & withheld & withheld \\
\cmidrule{1-8}
OLMo & all admissible & 13 & 108,099 & 0.283 & 0.423 & 0.460 & 0.419 \\
\midrule
Mistral & Kuhn & 4 & 9,041 & 0.23 & 0.46 & 0.53 & 0.36 \\
Mistral & Leduc & 4 & 51,208 & 0.31 & 0.40 & 0.37 & 0.38 \\
Mistral & Goofspiel & 1 & 20,000 & 0.66 & 0.57 & 0.62 & 0.95 \\
Mistral & Liar's Dice & 4 & 16,975 & 0.22 & 0.37 & 0.43 & 0.43 \\
Mistral & Dark Hex & 1 & 9,702 & 0.36 & 0.47 & 0.33 & 0.17 \\
\cmidrule{1-8}
Mistral & all admissible & 14 & 106,926 & 0.288 & 0.428 & 0.450 & 0.417 \\
\bottomrule
\end{tabular}
\end{table}

\paragraph{Consistency against the equilibrium posterior.}
Table~\ref{tab:p7bpost} scores the logged action against the best response to the equilibrium posterior, $\mathrm{BR}(\mu)$, on the same rows as against the stated belief, $\mathrm{BR}(\hat b)$, each with its own chance level.
Pooled over the admissible cells the two margins over chance are alike: \OptPostStatedMarginQwen{} against \OptPostPosteriorMarginQwen{} for Qwen, \OptPostStatedMarginOlmo{} against \OptPostPosteriorMarginOlmo{} for OLMo and \OptPostStatedMarginMistral{} against \OptPostPosteriorMarginMistral{} for Mistral.
Underneath they diverge by game: in Kuhn the action agrees more with the stated belief than with the posterior, and in Dark Hex more with the posterior, where the chance level also rises.
The action therefore does not track something closer to the equilibrium posterior than the agent states; it tracks both about equally.

\begin{table}[!ht]
\centering\small
\setlength{\tabcolsep}{4pt}
\caption{\textbf{\pgb{} against the stated belief and against the equilibrium posterior.} Agreement of the logged action with the best-response set, the chance level of that set and the margin, for the stated belief and for the equilibrium posterior on one population, per (model, game) over the admissible cells weighted by rows.}
\label{tab:p7bpost}
\begin{tabular}{@{}llrrrrrrr@{}}
\toprule
Model & Game & Cells & P7b & Chance & Margin & P7b & Chance & Margin \\
 & & & \multicolumn{3}{c}{against BR($\hat b$)} & \multicolumn{3}{c}{against BR($\mu$)} \\
\midrule
Qwen & Kuhn & 4 & 0.63 & 0.50 & 0.13 & 0.52 & 0.50 & 0.02 \\
Qwen & Leduc & 4 & 0.62 & 0.47 & 0.15 & 0.66 & 0.47 & 0.19 \\
Qwen & Goofspiel & 1 & 0.65 & 0.42 & 0.24 & 0.65 & 0.36 & 0.29 \\
Qwen & Liar's Dice & 4 & 0.44 & 0.14 & 0.30 & 0.43 & 0.14 & 0.29 \\
Qwen & Dark Hex & 1 & 0.35 & 0.41 & $-$0.07 & 0.67 & 0.62 & 0.05 \\
\cmidrule{1-9}
Qwen & all admissible & 14 & 0.56 & 0.38 & 0.18 & 0.58 & 0.39 & 0.19 \\
\midrule
OLMo & Kuhn & 4 & 0.52 & 0.50 & 0.02 & 0.57 & 0.50 & 0.07 \\
OLMo & Leduc & 4 & 0.58 & 0.46 & 0.12 & 0.41 & 0.46 & $-$0.05 \\
OLMo & Goofspiel & 1 & 0.00 & 0.36 & $-$0.36 & 0.00 & 0.36 & $-$0.36 \\
OLMo & Liar's Dice & 4 & 0.09 & 0.13 & $-$0.04 & 0.20 & 0.13 & 0.07 \\
OLMo & Dark Hex & 0 & withheld & & & withheld & & \\
\cmidrule{1-9}
OLMo & all admissible & 13 & 0.35 & 0.36 & $-$0.01 & 0.33 & 0.36 & $-$0.03 \\
\midrule
Mistral & Kuhn & 4 & 0.67 & 0.50 & 0.17 & 0.48 & 0.50 & $-$0.02 \\
Mistral & Leduc & 4 & 0.55 & 0.48 & 0.06 & 0.65 & 0.48 & 0.16 \\
Mistral & Goofspiel & 1 & 0.57 & 0.49 & 0.08 & 0.65 & 0.36 & 0.28 \\
Mistral & Liar's Dice & 4 & 0.18 & 0.11 & 0.07 & 0.18 & 0.11 & 0.07 \\
Mistral & Dark Hex & 1 & 0.24 & 0.39 & $-$0.16 & 0.49 & 0.58 & $-$0.09 \\
\cmidrule{1-9}
Mistral & all admissible & 14 & 0.46 & 0.39 & 0.07 & 0.49 & 0.39 & 0.10 \\
\bottomrule
\end{tabular}
\end{table}

\paragraph{Consistency by belief accuracy.}
Table~\ref{tab:p7bterc} bins the information sets of each (model, game) by their mean \pga{} into row-weighted thirds and reads \pgb{} in each; binning is by information set because the scorer reports both estimands per information set and not per row.
No consistent relation appears: Qwen's consistency rises with belief error in Kuhn and falls in Leduc, Mistral's rises in Leduc, and the per-model rows mix games whose \pga{} scales differ, so they are not read.
Actions do not follow the stated belief more often where that belief is closer to the equilibrium posterior.

\begin{table}[!ht]
\centering\small
\setlength{\tabcolsep}{4pt}
\caption{\textbf{\pgb{} by belief accuracy.} Information sets of the admissible cells sorted by their mean \pga{} and cut into row-weighted thirds (low, mid, high error; the two cut points are the \pga{} values at the bin edges), with the row-weighted \pgb{} in each third.}
\label{tab:p7bterc}
\begin{tabular}{@{}llrrrrrr@{}}
\toprule
Model & Game & Cells & P7a cuts & P7b, low & P7b, mid & P7b, high & Rows \\
\midrule
Qwen & Kuhn & 4 & 0.00 / 0.25 & 0.32 & 0.76 & 0.80 & 44,167 \\
Qwen & Leduc & 4 & 0.15 / 0.28 & 0.91 & 0.48 & 0.48 & 94,638 \\
Qwen & Goofspiel & 1 & 0.62 / 0.69 & 0.60 & 0.51 & 0.86 & 30,000 \\
Qwen & Liar's Dice & 4 & 0.12 / 0.24 & 0.34 & 0.40 & 0.57 & 60,448 \\
Qwen & Dark Hex & 1 & 0.33 / 0.41 & 0.09 & 0.90 & 0.06 & 15,311 \\
\cmidrule{1-8}
Qwen & all admissible & 14 & 0.16 / 0.35 & 0.59 & 0.53 & 0.58 & 244,564 \\
\midrule
OLMo & Kuhn & 4 & 0.05 / 0.35 & 0.63 & 0.47 & 0.46 & 51,132 \\
OLMo & Leduc & 4 & 0.15 / 0.33 & 0.62 & 0.37 & 0.73 & 86,353 \\
OLMo & Goofspiel & 1 & 0.45 / 0.52 & 0.00 & 0.00 & 0.00 & 30,000 \\
OLMo & Liar's Dice & 4 & 0.20 / 0.36 & 0.14 & 0.10 & 0.02 & 70,614 \\
OLMo & Dark Hex & 0 & withheld & & & & \\
\cmidrule{1-8}
OLMo & all admissible & 13 & 0.17 / 0.38 & 0.51 & 0.24 & 0.30 & 238,099 \\
\midrule
Mistral & Kuhn & 4 & 0.00 / 0.35 & 0.67 & 0.81 & 0.52 & 49,041 \\
Mistral & Leduc & 4 & 0.25 / 0.34 & 0.28 & 0.68 & 0.69 & 91,208 \\
Mistral & Goofspiel & 1 & 0.60 / 0.71 & 0.43 & 0.50 & 0.82 & 29,999 \\
Mistral & Liar's Dice & 4 & 0.00 / 0.30 & 0.17 & 0.12 & 0.24 & 56,975 \\
Mistral & Dark Hex & 1 & 0.25 / 0.43 & 0.00 & 0.00 & 0.76 & 19,702 \\
\cmidrule{1-8}
Mistral & all admissible & 14 & 0.24 / 0.40 & 0.36 & 0.48 & 0.56 & 246,925 \\
\bottomrule
\end{tabular}
\end{table}

\paragraph{Invariance under option reordering, with a floor agent.}
The relabelling probe of \pa{} renames the action vocabulary and keeps the order of the listed options, so no trivial agent moves under it and \pa{} has no floor (Fig.~\ref{fig:verdicts}).
A second probe reverses the order of the listed options against a canonical twin collected at the same temperature, seeds and hands: M0, the first 100 hands of seeds 42 to 46, every game and opponent.
It was collected as a diagnostic and is read here as a property.
Under reordering a position-following agent, one that always takes the first listed option, changes its action at every paired node.
Its paired distance is 1 and it is the floor; an agent that chooses by action sits at 0, the ceiling.
The canonical twin is the headline condition: its prompt specification hash equals the M0 arm's.
Its paired distance from the M0 arm at the same seed is \OrdTwinQwenAll{} for Qwen, \OrdTwinOlmoAll{} for OLMo and \OrdTwinMistralAll{} for Mistral, at most \OrdTwinMaxQwen{} in any game, within the null band of Fig.~\ref{fig:verdicts}.
Table~\ref{tab:p1order} gives the paired distance per game with its spread over seeds and the null band between canonical seeds.
It adds the three-way split of the diagnostic specified before the probe ran: whether the reversed arm's modal choice at a node is the same action, the same position, or neither.
Reordering moves \OrdTvQwenAll{} of Qwen's paired decisions, \OrdTvOlmoAll{} of OLMo's and \OrdTvMistralAll{} of Mistral's.
The three close \OrdAncQwenAll{}, \OrdAncOlmoAll{} and \OrdAncMistralAll{} of the gap from the position-following agent.
It stays out of Fig.~\ref{fig:verdicts} because uniform-random, constant-name and Bayesian play all reach its ceiling, the reason \pa{} under relabelling carries no floor agent there, so a coloured column would reward invariance that costs nothing.
The pooling weights games by paired information sets, as \pa{} is pooled. A cell at the floor means the reversal changed every choice, not that the choice followed position.
An agent that repeats one action name is invariant under reordering and sits at the ceiling; one that repeats a position changes its action at every node and sits at the floor.
The headline state-blind share cannot tell the two apart, which is what this probe adds, so no cell is withheld for state-blindness.
The split is not position-following: the same action survives at \OrdSameActionQwenAll{}, \OrdSameActionOlmoAll{} and \OrdSameActionMistralAll{} of informative nodes, the mirrored position at \OrdSamePositionQwenAll{}, \OrdSamePositionOlmoAll{} and \OrdSamePositionMistralAll{}.
Liar's Dice is the exception: reversing its list of bids leaves OLMo's choice neither the same action nor the same position at \OrdNeitherOlmoLiarsdice{} of nodes and Mistral's at \OrdNeitherMistralLiarsdice{}, so there the reversal scrambles the choice.
Mistral's Leduc choice follows the action at every one of its \OrdPairedMistralLeduc{} paired nodes, so its state-blind share of 1 is a constant action name, which reversal leaves unchanged.
The Dark Hex and Goofspiel rows rest on \OrdPairedSmallLo{} to \OrdPairedSmallHi{} paired nodes and carry no label in the diagnostic.

\begin{table}[!ht]
\centering\small
\setlength{\tabcolsep}{4pt}
\caption{\textbf{\pa{} under option reordering.} Paired total-variation distance between the canonical and the reversed-order arm, mean and spread over the five seeds, with the null band between canonical seeds; the position-following agent scores 1 and an agent that chooses by action 0. The last three columns are the diagnostic's shares of informative nodes at which the reversed arm's modal choice is the same action, the mirrored position, or neither.}
\label{tab:p1order}
\begin{tabular}{@{}llrrrrrrr@{}}
\toprule
Model & Game & Paired & TVD & Spread & Null band & Same action & Same position & Neither \\
\midrule
Qwen & Kuhn & 10 & 0.10 & 0.00 & 0.00 & 0.90 & 0.10 & 0.00 \\
Qwen & Leduc & 147 & 0.20 & 0.03 & 0.00 & 0.85 & 0.15 & 0.00 \\
Qwen & Goofspiel & 12 & 0.16 & 0.01 & 0.00 & 0.78 & 0.22 & 0.00 \\
Qwen & Liar's Dice & 49 & 0.23 & 0.04 & 0.02 & 0.76 & 0.06 & 0.18 \\
Qwen & Dark Hex & 2 & 1.00 & 0.00 & 0.00 & 0.00 & 0.50 & 0.50 \\
\cmidrule{1-9}
Qwen & all games & 221 & 0.21 & & 0.01 & 0.82 & 0.13 & 0.04 \\
\midrule
OLMo & Kuhn & 12 & 0.00 & 0.00 & 0.00 & 1.00 & 0.00 & 0.00 \\
OLMo & Leduc & 110 & 0.10 & 0.02 & 0.00 & 0.81 & 0.19 & 0.00 \\
OLMo & Goofspiel & 2 & 1.00 & 0.00 & 0.06 & 0.00 & 0.00 & 1.00 \\
OLMo & Liar's Dice & 35 & 0.84 & 0.03 & 0.01 & 0.16 & 0.08 & 0.76 \\
OLMo & Dark Hex & 2 & 1.00 & 0.00 & 0.02 & 0.00 & 0.00 & 1.00 \\
\cmidrule{1-9}
OLMo & all games & 161 & 0.27 & & 0.00 & 0.71 & 0.16 & 0.14 \\
\midrule
Mistral & Kuhn & 9 & 0.33 & 0.00 & 0.00 & 0.67 & 0.33 & 0.00 \\
Mistral & Leduc & 177 & 0.00 & 0.00 & 0.00 & 1.00 & 0.00 & 0.00 \\
Mistral & Goofspiel & 6 & 0.67 & 0.00 & 0.00 & 1.00 & 0.00 & 0.00 \\
Mistral & Liar's Dice & 38 & 0.93 & 0.04 & 0.00 & 0.08 & 0.03 & 0.90 \\
Mistral & Dark Hex & 2 & 1.00 & 0.00 & 0.00 & 0.00 & 1.00 & 0.00 \\
\cmidrule{1-9}
Mistral & all games & 232 & 0.19 & & 0.00 & 0.87 & 0.02 & 0.11 \\
\bottomrule
\end{tabular}
\end{table}

\paragraph{An adaptive agent for adaptivity: a negative result for the estimand.}
\pb{} is the shrinkage of the best-response gap from the first thousand hands to the last, and no collected arm said how much of that gap an agent that can adapt would close against these opponents.
We ran one through the same pipeline: a best responder to an empirical opponent model built from what the LM's memory also holds, the public actions of every finished hand and its own payoff, never the opponent's card.
The model counts the opponent's actions at each public prefix, smoothed towards the CFR+ reference with one pseudo-count.
The best response is recomputed every 50 hands, and nothing in its action reads a random number.
It plays the M1 arms' shape, 2000 hands and ten seeds against the four opponents in every game, and is scored by the unchanged \pb{} scorer.
Table~\ref{tab:p2ceiling} lists its two-bin delta beside the LM arms' and its mean reward in the first hundred hands and the rest.
Its two-bin delta runs from \PTwoRefLo{} to \PTwoRefHi{} and is within \PTwoRefNearTol{} of zero in \PTwoRefNear{} of \PTwoRefCells{} cells, inside the range of the LM arms' deltas (Table~\ref{tab:p2ceiling}), for two reasons.
Against the exploitable opponent it adapts inside the first bin.
Its reward against always-bet rises from \PTwoFirstKuhnAlwaysbet{} to \PTwoRestKuhnAlwaysbet{} in Kuhn, \PTwoFirstLiarsdiceAlwaysbet{} to \PTwoRestLiarsdiceAlwaysbet{} in Liar's Dice and \PTwoFirstGoofspielAlwaysbet{} to \PTwoRestGoofspielAlwaysbet{} in Goofspiel, and then stays level.
In Leduc against always-bet the delta has the wrong sign (\PTwoRefLeducAlwaysbet{}), as against CFR+ below.
Against the reference-like opponents there is little to exploit, and the public-prefix model cannot see what there is: pooled over the four opponents, the scorer's empirical table covers \PTwoCovLiarsdice{} of the information sets its best response visits in Liar's Dice and \PTwoCovGoofspiel{} in Goofspiel.
Under this estimand the ceiling therefore sits at the floor and \pb{} stays uncoloured in Fig.~\ref{fig:verdicts}.
Two things follow for the next run.
The trajectory must be fine enough to see adaptation inside the first hundred hands, which the two hand-index bins cannot.
And the public-prefix model is not the right ceiling agent against a card-conditioned opponent.
In Leduc against CFR+ its reward falls over hands (delta \PTwoRefLeducCfrref{}), since the model marginalises the opponent's card and the best response to it is exploitable.
The hindsight model keyed by the opponent's full information set needs a change to the instrument and belongs to the headline run.

\begin{table}[!ht]
\centering\small
\setlength{\tabcolsep}{3.5pt}
\caption{\textbf{The adaptive agent beside the LM arms on \pb{}.} Per (game, opponent) cell: the adaptive agent's mean reward over hands 1 to 100 and over hands 101 to 2000, its two-bin shrinkage delta (mean reward over hands 1 to 1000 minus mean reward over hands 1001 to 2000, negative where the gap shrank; mean over ten seeds), and the LM arms' deltas at M1 and M2, the per-cell values the grid pools.}
\label{tab:p2ceiling}
\begin{tabular}{@{}llrrrrrrrrr@{}}
\toprule
Game & Opponent & \multicolumn{3}{c}{Adaptive agent} & \multicolumn{2}{c}{Qwen} & \multicolumn{2}{c}{OLMo} & \multicolumn{2}{c}{Mistral} \\
\cmidrule(lr){3-5}\cmidrule(lr){6-7}\cmidrule(lr){8-9}\cmidrule(lr){10-11}
 & & hands 1--100 & 101--2000 & delta & M1 & M2 & M1 & M2 & M1 & M2 \\
\midrule
Kuhn & CFR+ & $-$0.04 & $-$0.04 & 0.00 & $-$0.02 & $-$0.04 & $-$0.00 & $-$0.01 & $-$0.02 & $-$0.02 \\
Kuhn & Bayesian & 0.16 & 0.26 & $-$0.02 & $-$0.01 & $-$0.05 & $-$0.01 & $-$0.02 & $-$0.02 & $-$0.03 \\
Kuhn & RL self-play & $-$0.05 & $-$0.06 & $-$0.00 & $-$0.03 & $-$0.05 & $-$0.00 & $-$0.01 & $-$0.02 & $-$0.03 \\
Kuhn & Always-bet & 0.14 & 0.34 & $-$0.04 & 0.01 & $-$0.06 & 0.00 & 0.00 & $-$0.02 & $-$0.03 \\
\midrule
Leduc & CFR+ & $-$0.09 & $-$0.26 & 0.16 & 0.02 & 0.00 & 0.09 & 0.13 & 0.06 & 0.09 \\
Leduc & Bayesian & 0.55 & 1.04 & $-$0.00 & 0.05 & 0.02 & 0.10 & $-$0.04 & 0.06 & 0.08 \\
Leduc & RL self-play & $-$0.18 & $-$0.49 & 0.01 & 0.07 & 0.06 & 0.28 & $-$0.09 & 0.08 & 0.10 \\
Leduc & Always-bet & 1.23 & 1.29 & 0.07 & 0.04 & 0.04 & 0.01 & 0.01 & 0.10 & 0.08 \\
\midrule
Goofspiel & CFR+ & 0.00 & 0.00 & 0.00 & $-$0.00 & 0.02 & 0.02 & 0.01 & $-$0.10 & $-$0.00 \\
Goofspiel & Bayesian & 0.00 & 0.00 & 0.00 & $-$0.00 & 0.04 & 0.00 & $-$0.00 & 0.00 & 0.00 \\
Goofspiel & RL self-play & 0.01 & 0.01 & 0.00 & $-$0.01 & 0.02 & 0.01 & 0.01 & $-$0.17 & 0.02 \\
Goofspiel & Always-bet & 2.12 & 2.50 & $-$0.04 & 0.02 & 0.21 & $-$0.14 & 0.00 & $-$0.08 & 0.00 \\
\midrule
Liar's Dice & CFR+ & $-$0.01 & $-$0.05 & $-$0.03 & $-$0.02 & $-$0.01 & 0.01 & $-$0.01 & $-$0.01 & $-$0.02 \\
Liar's Dice & Bayesian & 0.02 & 0.03 & $-$0.02 & 0.02 & 0.03 & $-$0.01 & $-$0.01 & 0.01 & 0.00 \\
Liar's Dice & RL self-play & $-$0.01 & $-$0.02 & $-$0.02 & $-$0.01 & 0.00 & 0.00 & $-$0.03 & $-$0.01 & $-$0.01 \\
Liar's Dice & Always-bet & 0.64 & 0.84 & $-$0.02 & $-$0.05 & 0.01 & 0.01 & 0.00 & 0.02 & $-$0.05 \\
\midrule
Dark Hex & CFR+ & 0.00 & 0.00 & 0.00 & $-$0.09 & 0.00 & 0.02 & 0.15 & 0.02 & 0.12 \\
Dark Hex & Bayesian & 0.00 & 0.00 & 0.00 & $-$0.00 & 0.04 & 0.00 & 0.00 & 0.00 & 0.00 \\
Dark Hex & RL self-play & 0.00 & 0.00 & 0.00 & $-$0.81 & 0.03 & $-$0.00 & 0.03 & 0.01 & 0.01 \\
Dark Hex & Always-bet & 1.00 & 1.00 & 0.00 & $-$0.84 & 0.06 & 0.00 & 0.00 & 0.00 & 0.00 \\
\bottomrule
\end{tabular}
\end{table}

\clearpage
\section{Glossary}
\label{app:glossary}
\begin{table}[h]
\centering\small
\setlength{\tabcolsep}{4pt}
\caption{\textbf{Terms used in this paper.}}
\label{tab:glossary}
\begin{tabular}{@{}p{0.22\linewidth}p{0.73\linewidth}@{}}
\toprule
Term & Meaning \\
\midrule
Node & one decision of an LM agent, logged as one row. \\
Mask & the set of actions legal at a node. \\
State-blind share & the share of a cell's nodes at which the action is the same across prompts that differ in state (Section~\ref{sec:stateblind}). \\
Constant-action share & the share of nodes whose action is the same across hands with the same mask, pooled over seats (Section~\ref{sec:memory}). \\
Seat-only cell & a cell in which no seat saw two distinct prompts; it carries no claim. \\
Stated belief & the belief an LM states about the opponent's hidden state, in a separate call at the same node. \\
Equilibrium posterior & the Bayes posterior over the opponent's hidden state when the opponent plays the CFR+ policy. \\
Best-response set & the actions that attain the best expected value under a stated belief. \\
Chance level & the probability that a uniformly random action lies in the best-response set (Appendix~\ref{app:chance}). \\
Equal-value share & the share of nodes whose chosen action is within 0.01 of the best action under the CFR+ reference's value function. \\
Admissibility gate & the rule that decides whether a cell's belief score may be taken as evidence about the model (Section~\ref{sec:instrument}). \\
Floor agent & a trivial agent or rule scored on the same rows or through the same pipeline (uniform-random player, uniform stater, chance level); a property's reading is the fraction of the gap from it to the ceiling that the LM closes (Fig.~\ref{fig:verdicts}). \\
Bayesian agent & the ceiling agent: it states the equilibrium posterior and acts on it; the reference serves as ceiling where Table~\ref{tab:properties} says so. \\
Withheld cell & a cell whose value is shown and whose score is withheld, with the reason structural, thin or failed. \\
Defect rates & five measured rates printed beside each belief score: null belief, zero posterior, key-set mismatch, support mismatch and identity (Appendix~\ref{app:defects}). \\
Arm & one LM agent over all five games and four opponents, at one memory level and temperature. \\
Ladder & one (model, game) pair read at M0, M1 and M2 (Appendix~\ref{app:ladders}). \\
M0, M1, M2 & memory levels: none, a running summary line, and the summary plus a transcript of recent hands. \\
Uniform-random player & a player choosing uniformly among legal actions, run through the same pipeline as the floor for P3 and P6. \\
Instrument & the logged rows, the scorers and the gate. \\
TVD & total variation distance (Eq.~\ref{eq:tvd}). \\
CFR+ reference & the average policy of counterfactual regret minimisation, a provable equilibrium approximation. \\
P1 to P8 & invariance, adaptivity, randomisation, payoff sensitivity, calibration, sequential rationality, belief error and belief--action consistency (the two estimands P7a and P7b of one property), and value of information (Table~\ref{tab:properties}). \\
Driven node & an off-path node reached after scripted opening moves into a subgame the CFR+ policy almost never visits; the LM then plays on against CFR+. Used for P6. \\
Probe run & a paired replay of the main arm under a relabelling or a payoff rescaling, used for P1 and P4. \\
State-sensitivity check & the check that withholds a belief cell where the action stays constant across prompts a seat could tell apart while the stated belief also stays the same across them. \\
Exact basis & scores the probability mass the LM's own distribution puts on a target action set. \\
Sampled basis & scores the action the LM actually took. \\
\bottomrule
\end{tabular}
\end{table}
\end{document}